\documentclass{article}
\usepackage{ijcai26}
\usepackage{pdfpages}
\usepackage{times}
\usepackage{soul} 
\usepackage{url}
\usepackage[hidelinks]{hyperref}
\usepackage[utf8]{inputenc}
\usepackage[small]{caption}
\usepackage{graphicx}
\usepackage{amsmath}
\usepackage{amsthm}
\usepackage{booktabs}
\usepackage{algorithm}
\usepackage{algorithmic}
\usepackage[switch]{lineno}

\usepackage{pifont}
\newcommand{\cmark}{\ding{51}} % ✓
\newcommand{\xmark}{\ding{55}} % ✗

\usepackage{multirow}
\usepackage{adjustbox}
\usepackage{array}      % for m{} column
\usepackage{adjustbox}  % optional, for auto fit
\usepackage[table]{xcolor}
\definecolor{ClosedBG}{RGB}{245,245,245} % light gray
\definecolor{OpenBG}{RGB}{236,248,255}   % light blue
\usepackage{makecell}

\title{UniPCB: A Unified Vision-Language Benchmark for \\ Open-Ended PCB Quality Inspection}
\author{
 \textbf{Fuxiang	Sun}$^{13*}$ \and 
 \textbf{Xi Jiang}$^{2}$\thanks{Contributed Equally.} \and
 \textbf{Jiansheng Wu}$^{3}$ \and 
 \textbf{Haigang	Zhang}$^{1\dagger}$ \and 
 \textbf{Feng Zheng}$^{2}$\thanks{Corresponding authors.} \and
 \textbf{Jinfeng	Yang}$^{1}$
 \vspace{0mm} \\
 {\normalfont
 $^{1}$Shenzhen Polytechnic University \quad
 $^{2}$Southern University of Science and Technology \\
 $^{3}$University of Science and Technology Liaoning 
 \vspace{0mm} \\
 \texttt{sunfuxiang@mail.szpu.edu.cn, jiangx2020@mail.sustech.edu.cn, wujiansheng@ustl.edu.cn, jfyang@szpu.edu.cn, } \\
 \texttt{zhanghg@szpu.edu.cn, f.zheng@ieee.org} 
 }
}

\begin{document}

\maketitle

\begin{abstract}
%多模态大型语言模型（MLLM）在工业质量检测领域展现出巨大潜力，但在印刷电路板（PCB）分析等复杂场景中仍显不足。通用模型往往难以解析高分辨率PCB图像中的密集布局与微小缺陷——这类任务需要特定领域的专业知识。我们认为，这种局限性不仅源于数据稀缺，更根本上源于数据集的碎片化及统一标准的缺失。
% 不只是数据匮乏  
% Multimodal Large Language Models (MLLMs) show promise for industrial quality inspection, but fall short in complex scenarios like Printed Circuit Board (PCB). General-purpose models often struggle to parse the dense layouts and minute defects in high-resolution PCB imagery—tasks requiring specific domain expertise. We contend that this limitation stems not merely from data scarcity, but fundamentally from dataset fragmentation and the absence of unified standards.
Multimodal Large Language Models (MLLMs) show promise for general industrial quality inspection, but fall short in complex scenarios, such as Printed Circuit Board (PCB) inspection. 
PCB inspection poses unique challenges due to densely packed components, complex wiring structures, and subtle defect patterns that require specialized domain expertise.
However, a high-quality, unified vision–language benchmark for quantitatively evaluating MLLMs across PCB inspection tasks remains absent, stemming not only from limited data availability but also from fragmented datasets and inconsistent standardization. 
% 为解决这一问题，我们推出UniPCB——首个面向开放式PCB质量检测的统一视觉语言基准，以及专用对话助手PCB-GPT。UniPCB通过系统化数据管道整合并标准化多源数据，构建包含5000张图像及6000组多模态问答对的基准测试集，覆盖三类标注场景。PCB-GPT基于该管道生成的指令数据集进行训练，采用模拟人类专家学习路径的渐进式课程体系。
To fill this gap, we propose UniPCB, the first unified vision-language benchmark for open-ended PCB quality inspection. UniPCB is built via a systematic pipeline that curates and standardizes data from disparate sources across three annotated scenarios. 
Furthermore, we introduce PCB-GPT, an MLLM trained on a new instruction dataset generated by this pipeline, utilizing a novel progressive curriculum that mimics the learning process of human experts. 
% To address this issue, we proposed an automatic pipeline consisting of. On this way,
%在UniPCB基准测试中，现有超大规模语言模型（MLLMs）在领域特定任务上表现欠佳，而PCB-GPT则树立了全新基准。尤其值得注意的是，其在精细缺陷定位方面的性能较最强竞争对手提升逾倍，在定位与分析能力上具有显著优势。我们将公开指令数据集、基准测试及模型，以推动工业多模态学习领域的未来研究。
Evaluations on the UniPCB benchmark show that while existing MLLMs falter on domain-specific tasks, PCB-GPT establishes a new baseline. Notably, it more than doubles the performance on fine-grained defect localization compared to the strongest competitors, with significant advantages in localization and analysis. We will release the instruction data, benchmark, and model to facilitate future research.
% in industrial multimodal learning.
\end{abstract}
\begin{figure}[!htbp]
    \centering
    \includegraphics[width=1\linewidth]{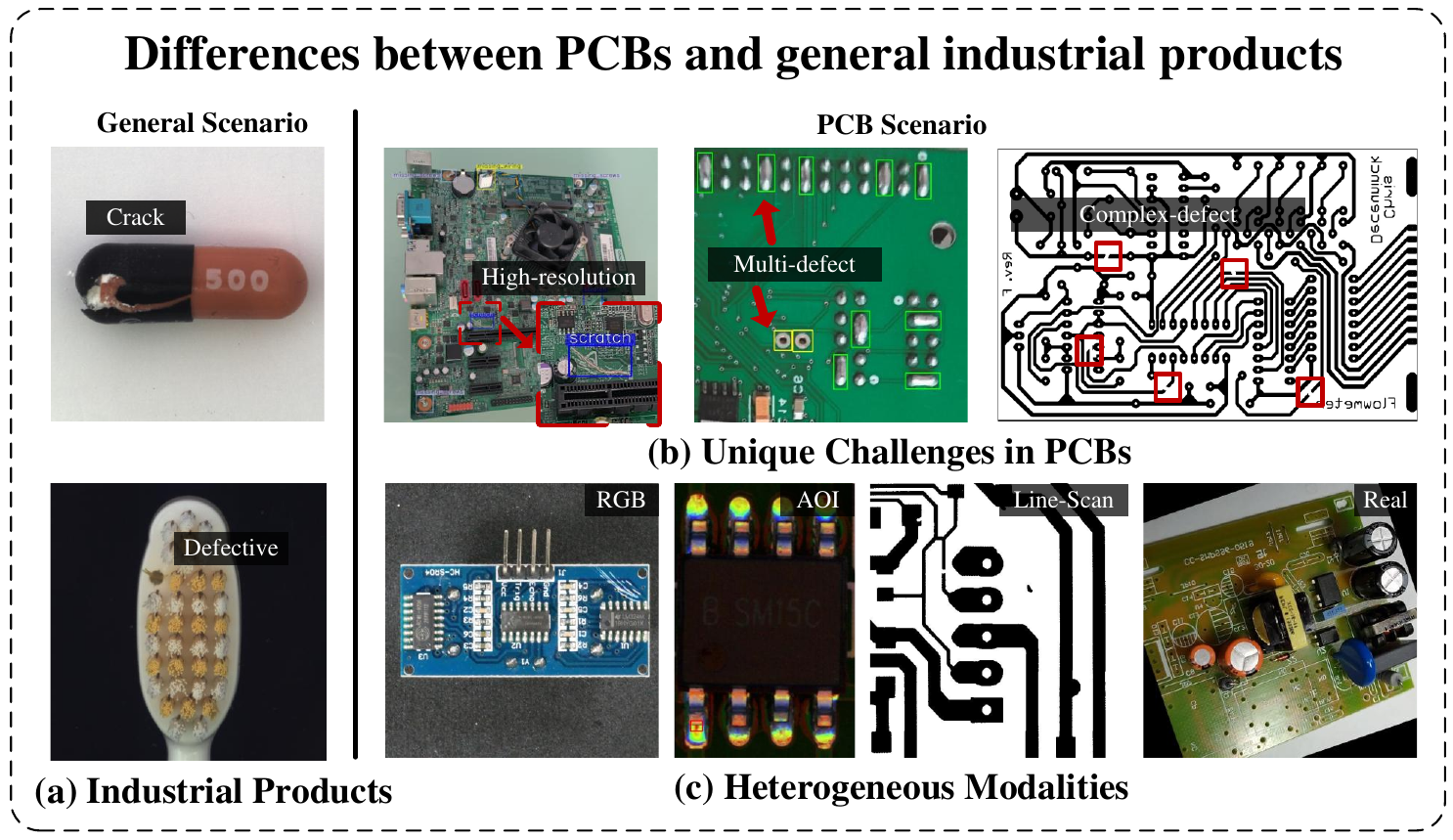} 
    \caption{Comparison between general industrial products and PCB inspection. (a) General products with obvious surface defects. (b) Unique PCB challenges: dense patterns, defect co-occurrence, and subtle cues. (c) Diverse imaging modalities are required to capture different defect characteristics.}
    \label{fig1}
\end{figure}
\section{Introduction} 
%在电子信息领域中，PCB作为至关重要的电子元件，直接影响着产品的性能与可靠性。与传统工业异常检测场景不同的是，PCB质量检测面临着更加复杂的挑战，其密集的几何结构、重复的纹理以及对精度的要求，并且缺陷往往表现为少量像素级扰动，容易被背景细节掩盖，还极易与正常走线、焊盘等设计结构混淆[引用]。
In the field of electronic information engineering, PCBs are fundamental components that directly determine the performance and reliability of electronic products. Unlike conventional industrial anomaly detection (IAD) scenarios, PCB quality inspection poses more complex challenges due to its highly dense geometric layouts, repetitive patterns, and strict precision requirements. Moreover, defects on PCBs often appear as subtle, pixel-level variations, which can be easily obscured by background details and are highly prone to being confused with normal design structures such as traces and solder pads~\cite{ling2023printed,sankar2022review}.

%尽管通用多模态大型语言模型近期取得了显著进展，例如Qwen-VL系列\cite{qwen2.5}、InternVL\cite{internvl3.5}和LLaVA\cite{llava-ov}在视觉理解方面展现出卓越能力，但应用于高度专业化的工业检测任务时往往表现欠佳，缺乏PCB质检所需的领域知识，难以区分正常设计与实际制造缺陷。同样地，像AnomalyGPT这类开创性IAD模型也仅限于简单的分类任务，难以应对PCB缺陷的高复杂性。尽管如此，从任务流程的角度看，多模态大模型（MLLM）仍具备贯穿PCB质检全过程的潜力——从缺陷的定位与识别，到后续的分析和风险评估。同时，其能够为最终的结论与决策提供可解释的推理依据。
Although the image understanding capabilities of general-purpose MLLMs, such as Qwen-VL series~\cite{qwen2,qwen2.5,qwen3}, InternVL series~\cite{internvl3,internvl3.5}, and LLaVA series~\cite{llava-ov,llava-next}, have steadily improved, their performance remains suboptimal when applied to highly specialized industrial scenarios like PCB quality inspection. Current general models lack the domain-specific knowledge required for PCB quality inspection, making it difficult for them to reliably distinguish normal circuit designs from genuine manufacturing defects. Similarly, pioneering IAD approaches such as AnomalyGPT~\cite{anomalygpt} are largely limited to coarse-grained classification and descriptive tasks, and struggle to handle the high complexity and fine-grained characteristics of PCB defects. Nevertheless, from a task-oriented perspective, MLLMs still have the potential to support the entire PCB quality inspection workflow. They can be applied to defect localization and identification, as well as subsequent analysis and risk assessment, with interpretable reasoning to aid final decision-making.
% . Moreover, MLLMs are capable of providing interpretable reasoning to justify inspection results and support final conclusions and decision-making processes.

%多语言大型模型在特定领域应用中的表现，本质上受限于其训练数据的质量与多样性。尽管先前研究将对有限数据集（如DeepPCB）的依赖归因于公开资源的匮乏，但我们认为主要障碍不仅在于数据稀缺，更在于其系统性混乱、严重碎片化、质量不一致以及完全缺乏统一标准。这种碎片化现象在众多数据集上显而易见——它们往往源自单一来源，覆盖缺陷范围狭窄，导致无法捕捉裸板（BPCB）向组装板（PCBA）演进过程中关键的分布变化。更棘手的是，一个更根本的问题——全行业缺乏统一的定义和标注标准——使得对分散的数据集进行公平比较几乎不可能，导致该领域缺乏统一基准来系统可靠地评估复杂检测模型。
The performance of MLLMs in domain-specific applications is fundamentally constrained by the quality and diversity of vision-language data. Although some prior work~\cite{pcb-deep} attributes the reliance on a limited number of datasets, such as DeepPCB~\cite{deeppcb}, to the scarcity of publicly available resources, we argue that the primary obstacle is not merely the scarcity of data, but rather its systemic disarray, severe fragmentation, inconsistent quality, and a complete lack of unified standards. This fragmentation is evident as many datasets are narrow in scope, often confined to a single modality (e.g., RGB images) and limited defect classes. Consequently, they fail to capture the critical distributional shifts that occur as bare printed circuit board (BPCB) evolve into assembled printed circuit board (PCBA). Compounding this, a more fundamental problem—the absence of industry-wide definitions and annotation standards—renders fair comparison across fragmented datasets nearly impossible, leaving the field without a unified benchmark for the systematic and reliable evaluation of complex inspection models.
%Compounding this, a more fundamental problem, the absence of industry-wide definitions and annotation standards, renders cross-dataset alignment nearly impossible, thereby undermining any attempt at systematic and reliable evaluation of complex inspection models.

%为应对这些挑战，我们提出UniPCB，一个专为PCB质量检测设计的大规模多场景视觉-语言基准数据集。有别于仅专注于单一检测任务的数据集，我们设计了一套系统化的数据构建流程。具体而言，我们收集互联网公开的PCB数据，覆盖多种工业成像模态，并与工程师协作严格定义了正常特征与缺陷标准，从而形成统一的标注规范与数据划分。在此基础上，我们将多模态指令微调方法[引用]引入PCB领域，构建面向PCB质检的对话式数据。该基准涵盖三类检测场景、十余种问题类型的多轮开放式问答数据集，包含6k张图像与23k组高质量多模态问答对。
To systematically study these challenges, we introduce UniPCB, a large-scale, multi-scenario vision–language benchmark tailored for PCB quality inspection. Unlike existing datasets that focus on a single inspection task, UniPCB is built through a systematic data construction pipeline. Specifically, we curate publicly available PCB data from the Internet, covering multiple industrial imaging modalities, and collaborate with experienced engineers to rigorously define normal characteristics and defect criteria, thereby establishing a unified annotation protocol and consistent data split. Building on this foundation, we further bring multimodal instruction tuning methods~\cite{zhang2023instruction,ouyang2022training} into the PCB domain to construct conversation-style data for PCB inspection. UniPCB spans three inspection scenarios and 14 question types in a multi-turn, open-ended question-answer(QA) setting, comprising 6k images and 23k high-quality multimodal question–answer pairs.

%基于 UniPCB 基准，我们对主流多语言模型[参考文献]进行了全面评估。实验表明，受限于领域知识的匮乏，现有通用MLLM仍存在明显的幻觉现象和缺陷定位不准确的问题。为弥补这一缺陷，我们提出了PCB-GPT——通过三阶段课程式训练的PCB质检视觉语言助手。通过模拟人类专家的认知学习路径，有效注入领域知识并保证结构化输出的稳定性。由此，PCB-GPT在多模态对话、精细缺陷识别及整体检测性能方面展现出显著优势。我们的贡献可概括如下：
Based on the UniPCB benchmark, we conduct a comprehensive evaluation of representative MLLMs~\cite{gemini2.5pro,qwen2.5,llava-next,internvl3.5,minicpmv4.5}. The results show that, due to limited domain knowledge, existing general-purpose MLLMs still suffer from noticeable hallucinations and inaccurate defect localization in PCB inspection. To mitigate these issues, we further propose PCB-GPT, a vision–language assistant for PCB quality inspection trained via a three-stage curriculum learning scheme. By mimicking the cognitive learning process of human experts, PCB-GPT effectively injects domain knowledge while improving the stability of structured outputs. As a result, PCB-GPT achieves advantages in multimodal dialogue, fine-grained defect identification, and overall inspection performance. Our contributions are summarized as follows:
%（1）我们提出了一套面向PCB质检的数据构建与质量控制管道，实现多源数据在训练和测评环节的标准化、清洗、划分等处理。
%（2）我们构建UniPCB——据我们所知，这是首个专为PCB质量检测设计的多模态、多任务基准数据集。该基准整合了BPCB与PCBA两个检测层级，涵盖三类标注场景及十四项质量检测子任务。
%（3）我们提出PCB-GPT——通过三阶段课程式训练的 PCB 质检视觉语言助手，支持从缺陷理解到风险评估的完整工作流程。
\begin{itemize}
    \item We propose a dedicated data construction and quality control pipeline for PCB inspection, enabling standardized processing of multi-source data, including cleaning, normalization, and dataset splitting.
    \item We construct UniPCB, to the best of our knowledge, the first multimodal, multi-task benchmark designed for PCB quality inspection. UniPCB unifies two inspection levels, BPCB and PCBA, and covers three annotated scenarios and fourteen quality inspection subtasks.
    \item We introduce PCB-GPT, a vision–language assistant for PCB quality inspection trained with a three-stage curriculum learning strategy, which supports the full workflow from defect understanding to risk assessment.
\end{itemize}
\begin{figure*}[!ht]
    \centering
    \includegraphics[width=1\textwidth,height=0.45\textheight,keepaspectratio]{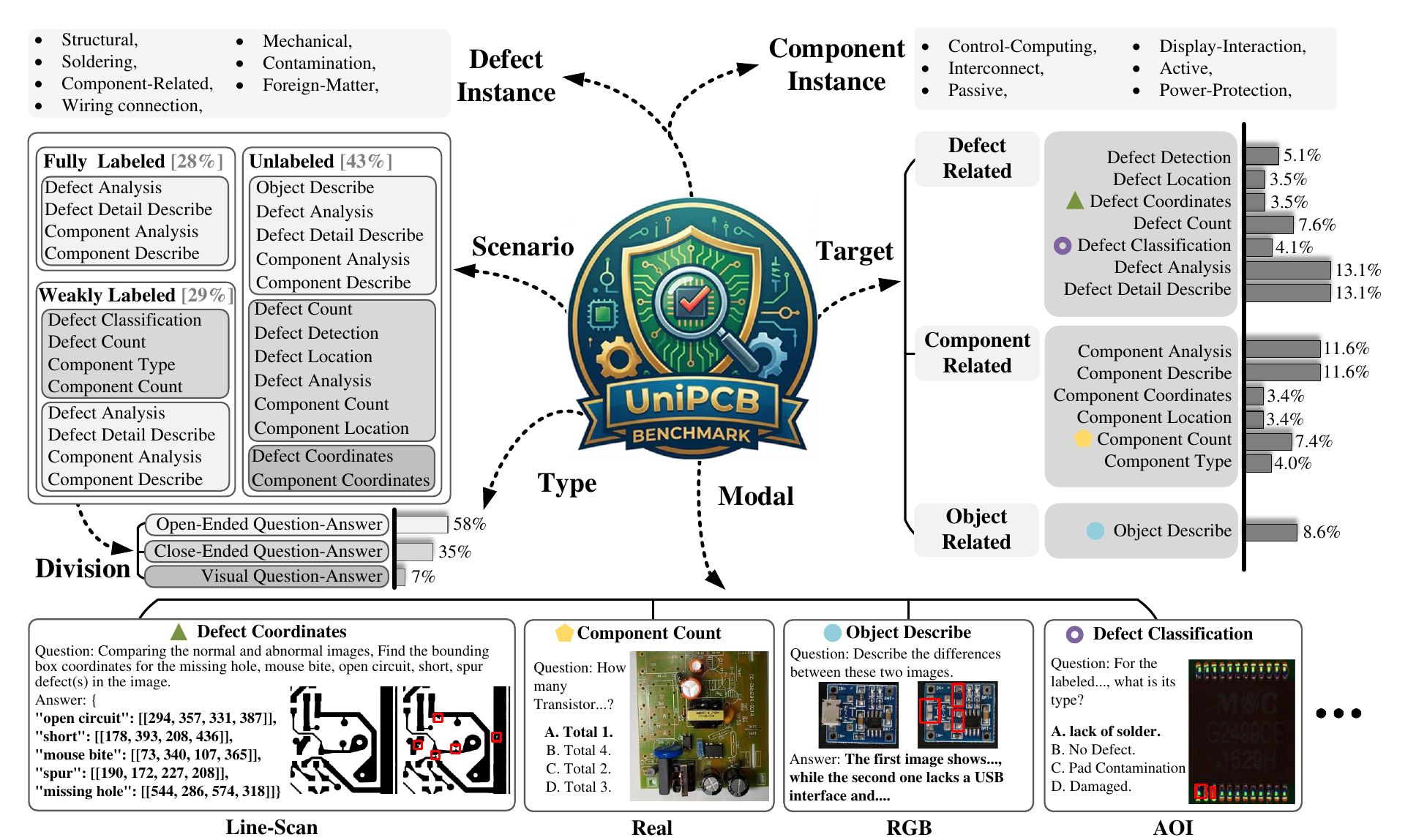} 
    \caption{Overview of UniPCB Benchmark. We summarize the three annotation scenarios, unified defect and component taxonomies, task by target type, and the overall task proportions, with representative QA examples shown at the bottom.}
    %, covering areas such as defect discrimination, object discrimination, component analysis, detection, classification, counting, location, and coordinates
    \label{fig2}
\end{figure*}
\section{Related Work}
\subsection{PCB Defect Analysis}
%PCB质量控制涵盖两个截然不同的制造阶段：BPCB和PCBA。前者针对开路等结构性异常，而后者则处理焊接错误等装配相关缺陷。尽管AOI和AXI等自动化系统能确保效率，但它们在遮挡和高元件密度场景下表现欠佳。深度学习方法（尤其是卷积神经网络）具备卓越的特征提取能力，但缺陷样本稀缺与标注标准不统一阻碍了其工业化部署。为突破这些数据层面的障碍并实现检测方法的严谨评估，我们提出基于开源PCB数据集构建的统一基准测试平台UniPCB。
As the backbone of electronic systems, the surface quality of PCB is critical to device reliability. Identifying anomalies, whether structural faults like open circuits or assembly errors like soldering defects, is therefore essential for yield optimization~\cite{BPCB&PCBA,bpcb-pcba}. Although traditional automated systems (AOI, AXI) ensure inspection efficiency~\cite{AOI&AXI,koblah2023comprehensive}, their reliability often drops in scenarios involving occlusion or high component density. Deep learning methods, notably CNN, provide stronger feature extraction capabilities yet face significant hurdles in industrial adoption due to data scarcity and inconsistent annotations~\cite{park2023analysis,chen2023comprehensive}. To overcome these data-centric barriers and enable the rigorous evaluation of detection methods, we introduce UniPCB, a unified benchmark constructed from open-source PCB datasets.
\subsection{MLLM}
%多模态大型语言模型（MLLMs）通常通过跨模态对齐将视觉编码器与大型语言模型（LLMs）相连接，从而实现字幕生成和视觉问答等功能}。该领域已分化为基于对比学习的检索（如CLIP）与指令对齐生成（如BLIP、LLaVA）两大方向。尽管近期研究将区域边界框和开放词汇检测融入指令调优以增强空间定位能力，通用型MLLM仍难以满足复杂工业检测的严苛要求，尤其在精准定位和专业领域知识方面存在短板。
MLLMs typically bridge visual encoders with LLMs via cross-modal alignment to enable capabilities like captioning and visual question answering~\cite{liu2023visual}. This field has bifurcated into contrastive learning for retrieval (e.g., CLIP~\cite{CLIP}) and instruction-aligned generation (e.g., BLIP~\cite{blip}, LLaVA~\cite{llava-ov}). While recent approaches have integrated region bounding boxes and open-vocabulary detection into instruction tuning to improve spatial grounding~\cite{peng2023kosmos,minderer2022simple}, general-purpose MLLMs generally fail to satisfy the stringent requirements of complex industrial inspection, particularly regarding precise localization and specialized domain knowledge.
\subsection{MLLM for Industrial Anomaly Detection}
%将超大规模语言模型引入异常检测已成为近期研究的焦点。该领域正从易于错误累积的级联设计转向统一的端到端框架。值得关注的方法包括采用群组相对策略优化（GRPO）提升推理能力的Anomaly-R1和OmniAD，以及强调精细缺陷感知能力的VMAD。与此同时，FabGPT等领域专用模型揭示了通用型MLLM在复杂工业任务中的局限性。然而，高质量指令语料库与统一评估框架的匮乏阻碍了技术进步。MMAD和EIAD等基准测试表明主流模型仍缺乏工业部署所需的精度。为此，我们针对PCB检测领域定制数据整理与训练策略，以弥合这些能力缺口。
Bringing MLLMs into IAD has become a focal point of recent research~\cite{survey}. The field is pivoting from cascaded designs~\cite{anomalygpt,myriad}—which are prone to error accumulation—toward unified end-to-end frameworks. Notable approaches include Anomaly-R1 and OmniAD~\cite{anomalyr1,ominiad,LR-IAD}, which employ Group Relative Policy Optimization (GRPO)~\cite{grpo} for improved reasoning, and VMAD~\cite{vamd,Traid}, which emphasizes fine-grained defect awareness. Meanwhile, domain-specific models like FabGPT~\cite{fabgpt} expose the inadequacy of generalist MLLMs for intricate industrial tasks. However, advancement is stalled by the paucity of high-quality instruction corpora and unified evaluation frameworks. Benchmarks such as MMAD~\cite{mmad} and EIAD~\cite{eiad} reveal that mainstream models still lack the precision required for industrial deployment. Consequently, we investigate data curation and training strategies for PCB inspection to bridge these capability gaps.
\section{UniPCB: Dataset and Benchmark}
\subsection{Challenges and Motivation}
%传统工业检测侧重于粗糙表面纹理，而PCB质量控制则面临独特的结构限制（图1）。此处的缺陷不仅是表面异常，更与电路图案紧密相连。这些缺陷往往极其微小且被密集背景所掩盖，难以与正常的制造差异区分开来——尤其考虑到走线与焊盘等电路基本元素在视觉上的相似性。异构成像管道的处理更使问题复杂化——从RGB到线扫描模式的多种模态导致巨大域差异。为严谨解决这些问题，我们构建了基于强健数据清洗策略的多源数据集。通过感知哈希消除冗余、剔除存在几何或光度失真的样本、并经人工验证实现标签标准化，为复杂工业场景的基准测试提供了统一基准。
% While conventional industrial inspection focuses on coarse surface textures, PCB quality control faces unique structural constraints (Fig.~\ref{fig1}). Defects here are not merely surface anomalies but are intricately tied to the circuit patterns. They are often minute and obscured by the dense background, making them difficult to isolate from legitimate manufacturing variations, especially given the visual similarity between circuit primitives like traces and pads. This difficulty is magnified when dealing with heterogeneous imaging pipelines, where modalities ranging from RGB to line-scan patterns impose large domain discrepancies. To rigorously address these issues, we curated a multi-source dataset underpinned by a robust cleaning strategy. By leveraging perceptual hashing to eliminate redundancy, discarding samples with geometric or photometric distortions, and standardizing labels through manual verification, we provide a consistent foundation for benchmarking in complex industrial scenarios.
Unlike general industrial inspection targeting coarse textures, PCB quality inspection involves unique structural constraints (Fig.~\ref{fig1}). Defects are often minute, and visually indistinguishable from normal circuit primitives (e.g., traces and pads), complicating detection. This challenge is compounded by heterogeneous imaging modalities, ranging from RGB to line-scan. To address this, we constructed a multi-source dataset using a strict cleaning pipeline. By implementing perceptual hashing deduplication, distortion filtering, and label standardization, we established a consistent benchmark for complex industrial scenarios. 
\begin{figure}[t]
    \centering
    \includegraphics[width=1\linewidth]{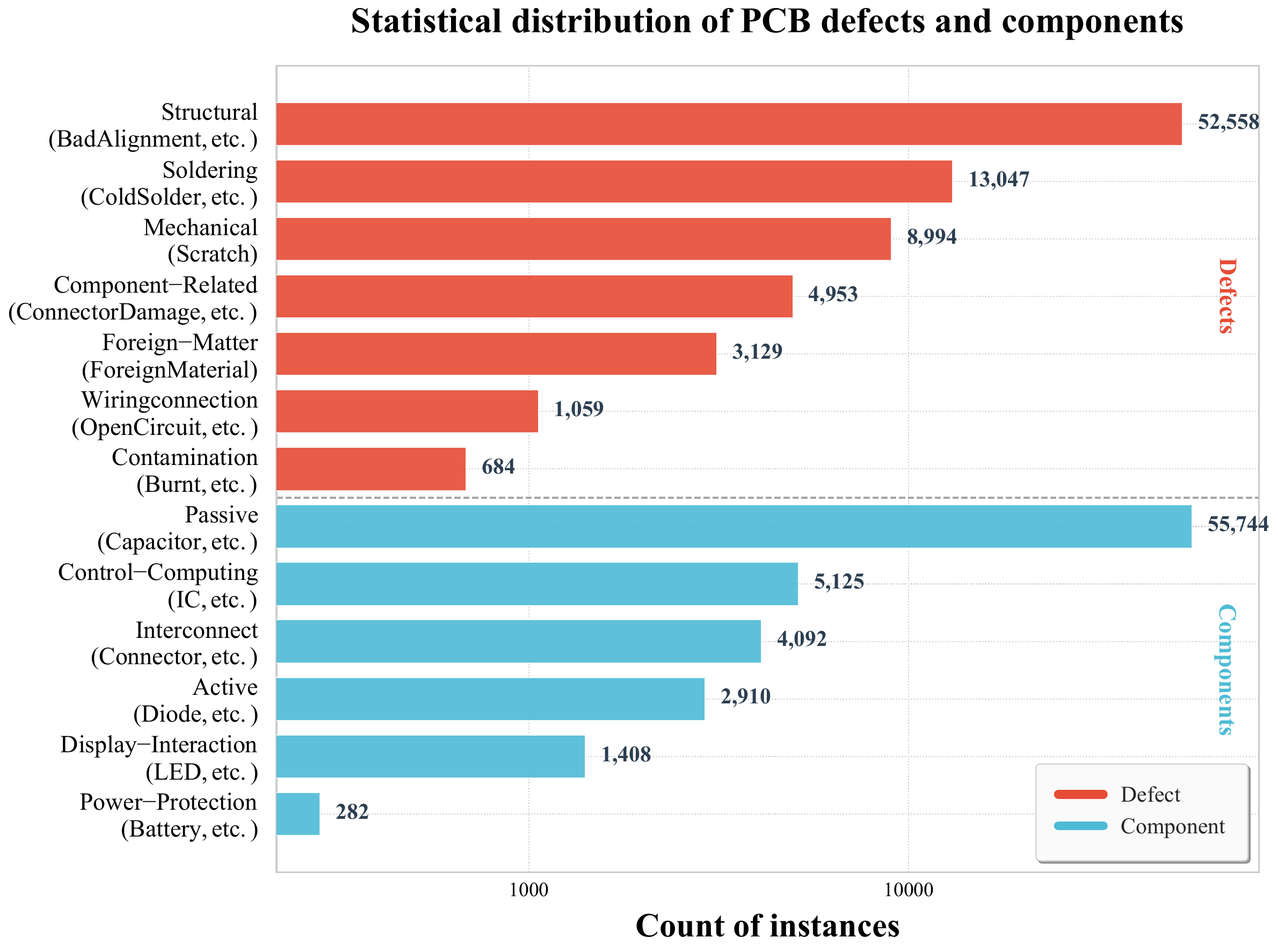}
    \caption{Statistical distribution of defect and component instances under the unified taxonomy.}
%We report the instance counts for seven defect super-categories (e.g., structural, soldering, component-related, wiring connection, mechanical, contamination, and foreign matter) and six component functional domains (e.g., control & computing, interconnect, passive, active, display & interaction, and power & protection).
    \label{fig4}
\end{figure}
\paragraph{Unified taxonomies}
%为了统一来自不同来源的标注信息，我们建立了一套统一的缺陷和组件分类系统。缺陷分为七大类（例如，结构缺陷、焊接缺陷、组件相关缺陷、布线连接缺陷、机械缺陷、污染、异物），组件则分为六个功能领域（例如，控制与计算、互连、无源器件、有源器件、显示/交互、电源与保护）。所有训练数据和基准标注都严格遵循这些分类系统。
To unify annotation information from different sources, we established a unified classification system for defects and components (Fig.~\ref{fig2}). Defects are categorized into seven categories (e.g., structural, soldering, component-related, wiring connection, mechanical, contamination, foreign-matter), and components are organized into six functional domains (e.g., control-computing, interconnect, passive, active, display-interaction, power-protection). Fig.~\ref{fig4} summarizes the category-wise instance distribution under this taxonomy. All training data and benchmark annotations consistently adhere to these classification systems.
\begin{figure*}[t]
    \centering
    \includegraphics[width=1\textwidth]{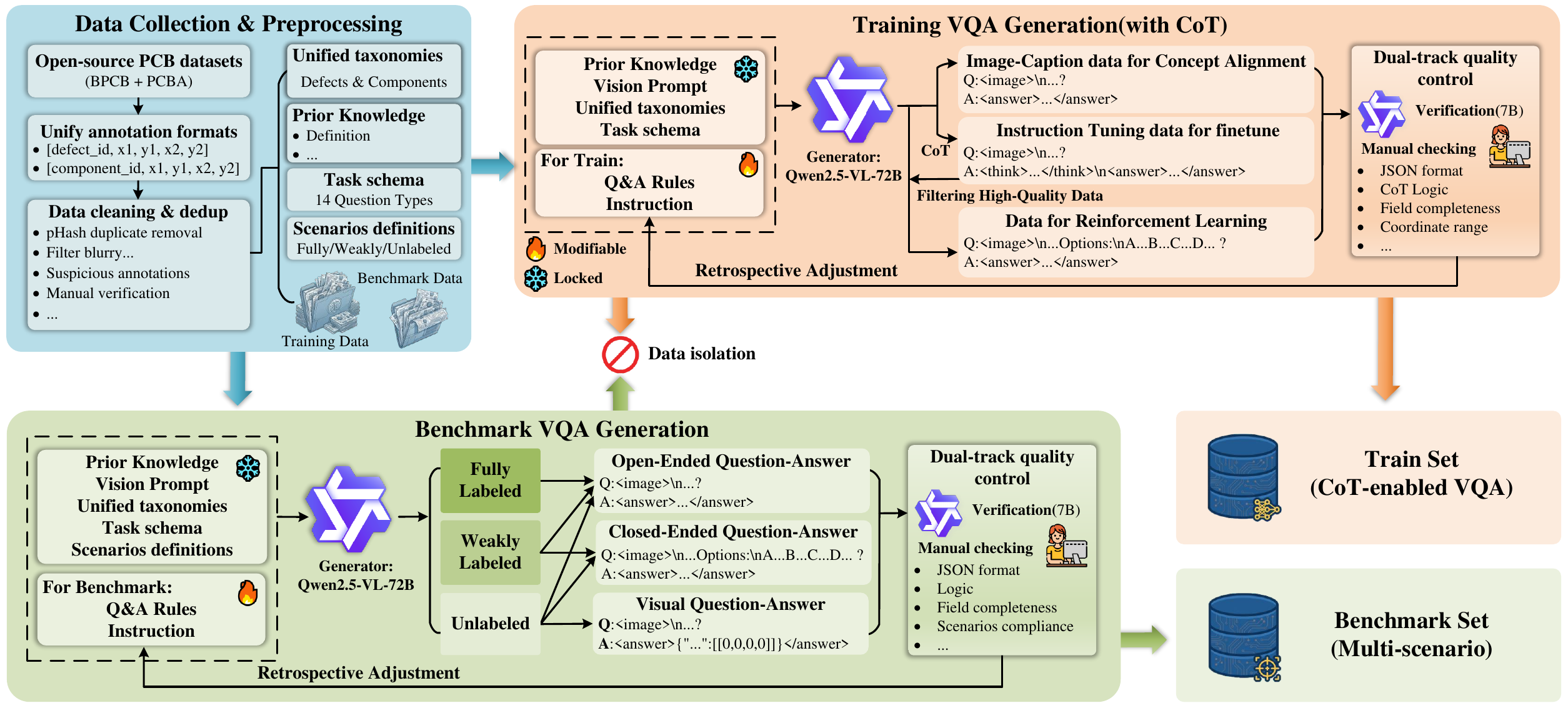}
    \caption{Data construction pipeline. We unify data sources, annotations, taxonomies, and a 14-type task schema, and construct two generation branches: a CoT-enabled training set and a multi-scenario benchmark. Each branch applies dual-track quality control with iterative prompt/rule refinement.}
    %The VQA data generation pipeline for UniPCB Benchmark. 
    \label{fig3}
\end{figure*}
\begin{figure}[t]
    \centering
    \includegraphics[width=1\linewidth]{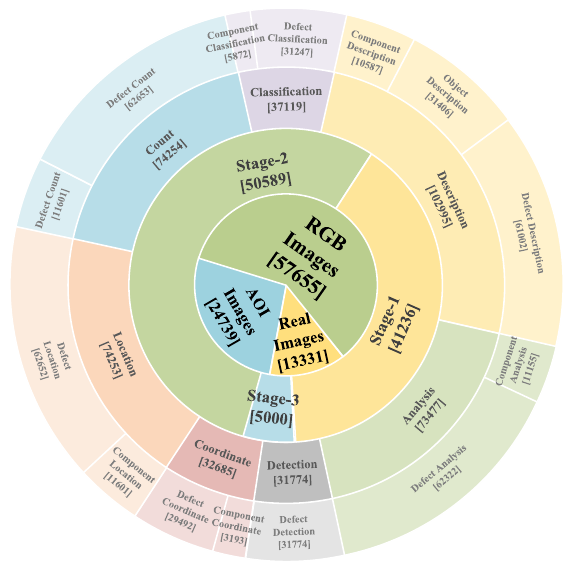}
    \caption{Rings from inner to outer represent imaging modalities (RGB/AOI/Real), data construction stages, and task families with fine-grained sample counts.}
%Overview of the UniPCB training data composition. Innermost ring denotes imaging modalities (RGB/AOI/Real), the middle ring indicates data construction stages, the outer rings show task families and fine-grained question types with their sample counts.
%UniPCB训练数据组成概览。最内环表示成像模态（RGB/AOI/实测），中间环表示数据构建阶段，外环显示任务族和细粒度问题类型及其样本数量。
    \label{fig5}
\end{figure}
\paragraph{Task and knowledge construction}
%在实际检查流程的指导下，我们定义了 14 个子任务，涵盖检测、计数、分类、定位、缺陷分析和复检建议。针对每种缺陷和组件类型，我们都创建了一个知识库条目，其中包含简短定义、常见原因、潜在影响和标准化表述。这种结构化的知识支持基于模板生成多样化且规范的问题和答案。
Guided by practical inspection workflows, we defined 14 subtasks covering detection, counting, classification, localization, defect analysis, and reinspection recommendations. For each defect and component type, we created a knowledge base entry that includes a short definition, common causes, potential impact, and standardized phrasing. This structured knowledge supports the generation of diverse yet controlled questions and answers based on templates.
\paragraph{VQA data generation and quality control}
%我们基于Qwen2.5-VL-72B-Instruct模型构建了视觉问答生成管道，将标注图像转化为多轮问答对。通过自适应模板机制，我们在不同模态间实现了标准化输出格式（如基于JSON坐标的定位信息）。为确保数据可靠性，我们实施了双轨质量控制机制：结合自动结构验证与分层人工审核，并利用反馈结果迭代优化提示词，以实现语义与结构的一致性。
We developed a VQA generation pipeline using Qwen2.5-VL-72B-Instruct to transform annotated images into multi-turn QA pairs. We utilized adaptive templates to enforce standardized output formats, such as JSON coordinates for localization, across diverse modalities. To ensure data reliability, we implemented a dual-track quality control mechanism combining automated structural validation with hierarchical human review, where feedback is used to iteratively refine prompts for semantic and structural consistency. 
\subsection{The UniPCB Benchmark}
\paragraph{Design Principles}
%UniPCB Benchmark基准测试旨在：(i) 模拟实际的检测工作流程，包含14个子任务。(ii) 尽可能要求输出结构化数据，以便进行自动化和可重复的性能评估。(iii) 有意包含一些具有挑战性的案例，例如小目标、严重遮挡、高度相似的组件和多缺陷场景。
The UniPCB Benchmark aims to (i) simulate practical inspection workflows, encompassing 14 subtasks. (ii) require structured outputs wherever possible to enable automated and repeatable performance evaluation. (iii) deliberately include challenging cases such as small targets, severe occlusions, highly similar components, and multi-defect scenes. 
\paragraph{Scenario definitions}
%为评估模型在不同标注方案下的鲁棒性，我们构建了三种场景。在完全标注场景中，每个对象均附有类别与边界框标注，用于评估模型在具备丰富先验信息时的分析能力。在弱标注场景中，仅提供边界框标注，要求模型基于外观和上下文推断类别与数量。在无标注场景中，完全不提供标签；提示语仅指定缺陷或目标部件，模型必须直接完成检测、定位与描述。针对部分困难样本，我们额外提供模板图像以辅助基于对比的缺陷检测。
To evaluate the model's robustness under different annotation schemes, we constructed three scenarios. In the fully labeled scenario, each query object is labeled with both category and bounding-box annotations to assess the model's analytical capabilities with rich prior information. The weakly labeled scenario only provides bounding-box annotations, requiring the model to infer categories and quantities from appearance and context. In the unlabeled scenario, no labels are provided. Instead, the prompt specifies the defect or component of interest, and the model must directly detect, localize, and describe it. %For more challenging samples, we provide a template image to aid in defect detection via comparison.
% 为保证评测公平性与结果可比性，三类场景在任务指令与生成模板上的设计上保持独立，并在样本抽取与扩充过程中严格实施隔离与去重，整体数据构建流程如图3.x所示。
To ensure fairness and comparability, the task instructions and template generation for the three scenarios are designed independently. Strict isolation and deduplication are applied throughout sample extraction and expansion. The overall data construction process is illustrated in Fig.~\ref{fig3}.
\paragraph{Evaluation metrics}
\label{sec:3.2.3}  
%该基准测试采用三种互补的评估协议。对于开放式问题（OQA），我们使用强大的大型语言模型（LLM）作为评估器。然后计算这五个分数的平均值，并将其归一化到[0,1]区间。此外，我们还使用BERTScore [aaaa]和Sentence-BERT相似度[aaaa]等常用自然语言生成指标作为补充参考。对于封闭式问题（CQA），包括单选题和多选题，我们通过将预测选项与正确答案进行匹配来计算准确率，并在必要时使用简单的启发式方法从自由文本中提取选项字母。
The benchmark employs three complementary evaluation protocols. For open-ended question-answer (OQA), we utilize a powerful LLM as an evaluator. The average of these five scores is then calculated and normalized to the [0,1] interval. 
% \begin{equation}
% \mathrm{Score} = \frac{\frac{1}{5}\sum_{i=1}^{5} S_i - S_{\min}}
% {S_{\max} - S_{\min}}, \quad \mathrm{Score} \in [0,1]
% \end{equation}
In addition, we employ common natural language generation metrics such as BERTScore~\cite{bertscore} and Sentence-BERT similarity~\cite{sentence-bert} as supplementary references. For closed-ended question-answer (CQA), including single-choice and multiple-choice questions, we calculate accuracy by matching predicted options against the correct answers, and when necessary, employ simple heuristics to extract option letters from free-form text. 
For visual question-answer (VQA), we match predicted boxes $P$ and ground-truth boxes $G$ using the Hungarian algorithm based on Intersection-over-Union (IoU) cost, considering a match valid if the IoU exceeds 0.3 for the same category. We then report Precision, Recall, and F1 scores.
%For visual question-answer (VQA), we match predicted boxes $P = \{p_i\}$ and ground-truth boxes $G = \{g_j\}$ using the Hungarian algorithm on the cost matrix
% \begin{equation}
% c_{ij} =
% \begin{cases}
% 1 - IoU(p_i, g_j), & \text{if } type(p_i) = type(g_j), \\
% 1, & \text{else}.
% \end{cases}
% \end{equation}
% where \begin{equation}
% IoU(p_i, g_j) = \frac{|p_i \cap g_j|}{|p_i \cup g_j|}
% \end{equation} 
% If the IoU exceeds the threshold $\tau=0.3$, the match is considered valid. We then calculate precision, recall, and F1 score based on true positives, false positives, and false negatives. 
All subsequent experiments follow this unified protocol.
\paragraph{Statistic Analysis}
%UniPCB数据集包含近6,000张图像和23,000组双语质量评估对，按三个渐进场景进行组织（图2）。值得注意的是，任务分布超越了简单感知范畴：基准测试中绝大部分内容聚焦于缺陷分析与细节描述。此设计迫使模型从单纯识别缺陷位置，转向推理缺陷成因及潜在影响，从而评估其工业理解深度而非仅限于表面检测能力。
% Overall, UniPCB constitutes a large-scale, bilingual evaluation platform comprising nearly 6,000 images and over 23,000 QA pairs. As visualized in Fig.~\ref{fig2}, the benchmark is structured into three progressive scenarios to rigorously test robustness under varying information density. Notably, the task distribution moves beyond simple perception: a dominant portion of the benchmark is dedicated to Defect Analysis and Detail Description. This design compels models to transition from merely identifying where a defect is to reasoning why it occurred and what it implies, thereby evaluating the depth of industrial understanding rather than just surface-level detection.
UniPCB features 6581 images and 23,359 bilingual QA pairs organized into three progressive scenarios. Notably, the task distribution moves beyond simple perception: a dominant portion of the benchmark is dedicated to Defect Analysis and Detail Description. This design compels models to transition from merely identifying where a defect is to reasoning why it occurred and what it implies, thereby evaluating the depth of industrial understanding rather than just surface-level detection.
\begin{table*}[!htbp]
\centering
\scriptsize
\setlength{\tabcolsep}{4pt}
\renewcommand{\arraystretch}{1.2}
% 建议去掉 height 限制，否则会强行压缩表格导致字体变形
\begin{adjustbox}{width=1\textwidth, center} 
\begin{tabular}{>{\centering\arraybackslash}m{1.4cm}|c|c|cc|ccc|ccc|c|c|c}
\toprule
\multirow{2}{*}{\rule{0pt}{2.6ex}Type} &
\multirow{2}{*}{\rule{0pt}{2.6ex}Model} &
\multirow{2}{*}{\rule{0pt}{2.6ex}Scale} &
\multicolumn{2}{c|}{Acc} &
\multicolumn{3}{c|}{S-BERT} &
\multicolumn{3}{c|}{BERTScore} &
\multirow{2}{*}{\rule{0pt}{2.6ex}F1} &
\multirow{2}{*}{\rule{0pt}{2.6ex}LLM as Judge} &
\multirow{2}{*}{\rule{0pt}{2.6ex}Average} \\
\cmidrule(lr){4-5}\cmidrule(lr){6-8}\cmidrule(lr){9-11}
& & &
P2 & P3 &
P1 & P2 & P3 &
P1 & P2 & P3 &
& & \\
\midrule
% =========================
% Closed-source block (3 rows)
% =========================
\rowcolor{ClosedBG}
& GPT5-Main       & -- & \underline{73.1} & \underline{66.1} & \textbf{77.0} & \textbf{71.2} & \textbf{72.2} & \textbf{70.5} & \textbf{71.0} & \textbf{70.5} & \underline{20.2} & \textbf{73.7} & \textbf{66.5} \\
\rowcolor{ClosedBG}
& Gemini2.5-Pro   & -- & \textbf{76.4} & \textbf{68.5} & \underline{71.9} & \underline{69.6} & \underline{66.5} & \underline{67.7} & \underline{69.3} & \underline{64.9} & \textbf{23.6} & \underline{70.8} & \underline{64.9} \\
\rowcolor{ClosedBG}
\multirow{-3}{*}{\cellcolor{ClosedBG}\makecell{Commercial\\MLLM}} % 移到最后一行，使用 -3
& Gemini2.5-Flash & -- & 63.4 & 46.5 & 70.9 & 67.4 & 63.6 & 67.6 & 68.5 & 63.7 & 17.0 & 69.1 & 59.8 \\
\midrule
% =========================
% IAD MLLM block (3 rows)
% =========================
\rowcolor{OpenBG}
& EMIT       & 8B & 36.4 & 48.5 & 69.3 & 60.0 & \textbf{70.2} & 69.2 & 67.1 & \textbf{70.0} & 10.5 & 60.2 & 56.1 \\
\rowcolor{OpenBG}
& IAD-R1     & 7B & 58.9 & 50.1 & 56.9 & 54.7 & 56.4 & 61.3 & 60.0 & 60.5 & 15.2 & 36.0 & 51.0 \\
\rowcolor{OpenBG}
\multirow{-3}{*}{\cellcolor{OpenBG}\makecell{IAD\\MLLM}} % 移到最后一行，使用 -3
& AnomalyGPT & 7B & 12.9 & 12.3 & 37.4 & 50.1 & 47.0 & 58.3 & 61.4 & 60.0 & --   & 34.6 & 37.4 \\
\midrule
% =========================
% Open Source block (11 rows)
% =========================
\rowcolor{OpenBG}
& DeepSeek2VL       & 16B & 17.5 & 24.4 & 57.8 & 47.3 & 53.3 & 64.6 & 62.2 & 63.0 & --   & 39.6 & 43.0 \\
\rowcolor{OpenBG}
& InternVL2.5       & 8B  & 56.7 & 39.0 & 66.0 & 64.0 & 66.6 & 68.1 & 67.7 & 65.6 & 14.0 & 58.7 & 56.6 \\
\rowcolor{OpenBG}
& InternVL3         & 8B  & 52.9 & 47.2 & 64.7 & 56.0 & 66.1 & 67.2 & 62.1 & 67.5 & 18.0 & 56.5 & 55.9 \\
\rowcolor{OpenBG}
& InternVL3.5       & 8B  & 57.5 & 43.5 & 65.5 & 53.7 & 64.5 & 67.7 & 61.1 & \underline{68.0} & 19.5 & 55.5 & 55.6 \\
\rowcolor{OpenBG}
& LLaVA-OV          & 8B  & 61.6 & \underline{55.4} & \underline{73.2} & 56.7 & \underline{68.7} & \underline{70.0} & 67.1 & 67.9 & -- & 59.8 & 58.1 \\
\rowcolor{OpenBG}
& MiMo-V2-Flash     & 15B & 40.0 & 47.7 & 42.0 & 64.0 & 45.4 & 57.8 & 67.0 & 61.4 & 16.1 & 42.6 & 48.4 \\
\rowcolor{OpenBG}
& MiniCPM-V4.5      & 8B  & 53.9 & 54.0 & 57.7 & \underline{66.3} & 65.6 & 61.0 & 67.9 & 67.2 & 15.5 & 58.2 & 56.7 \\
\rowcolor{OpenBG}
& Qwen2.5-VL        & 7B  & 53.2 & 44.2 & 66.7 & 50.4 & 63.7 & 68.1 & 60.7 & 66.5 & \underline{22.3} & 54.3 & 55.0 \\
\rowcolor{OpenBG}
& Qwen3-VL-Instruct & 8B  & 61.5 & 48.7 & 64.8 & 66.2 & 64.6 & 65.0 & \underline{68.2} & 64.1 & 22.2 & \underline{61.0} & \underline{58.6} \\
\rowcolor{OpenBG}
& Qwen3-VL-Think    & 8B  & \underline{65.6} & 54.6 & 64.9 & 65.5 & 58.0 & 65.6 & 66.5 & 65.0 & 20.1 & 58.7 & 58.5 \\
\rowcolor{OpenBG}
\multirow{-11}{*}{\cellcolor{OpenBG}\makecell{Open Source\\MLLM}} % 移到最后一行，使用 -11
& PCB-GPT (Ours)    & 7B  & \textbf{72.5} & \textbf{66.4} & \textbf{73.4} & \textbf{69.0} & 67.1 & \textbf{70.1} & \textbf{70.0} & 67.4 & \textbf{51.1} & \textbf{65.8} & \textbf{67.3} \\
\bottomrule
\end{tabular}
\end{adjustbox}
\caption{Quantitative comparison on the UniPCB Benchmark across Commercial, IAD, and Open-Source MLLMs. Bold and \underline{underlined} values indicate the best and second-best performance within each model category, respectively. P1, P2, and P3 denote the fully labeled, weakly labeled, and unlabeled settings.}
\label{table3}
\end{table*}
\subsection{Training Data}
\label{sec:3.2.4} 
% The constructed dataset is reorganized into three subsets. As illustrated in Fig.~\ref{fig5}, our training data span multiple imaging modalities and are organized by stages, task families, and fine-grained question types. the first stage utilized approximately 100k caption pairs sourced from both general datasets and PCB datasets, focusing on prominent and minimally distracting samples. This enabled the model to reliably associate visual patterns with standardized defect and component terminology. The second stage leverages roughly 300k multi-turn dialogues generated under the unified schema, covering diverse scenarios and question types while enforcing consistent output formats. The third stage uses a smaller but denser structured sample set of approximately 23k samples, where answers are strictly verifiable (e.g., MCQ, JSON coordinates), which enables reward-based optimization. Table~\ref{table1} summarizes the datasets contributing to each stage and the overall data volume. All stages share the same taxonomies and annotation conventions to minimize distribution shifts.
%如图~\ref{fig5}所示，构建的数据集按训练阶段、任务类别及精细化问题类型分层划分为三个子集，覆盖多种成像模态。第一阶段包含约10万对通用领域与PCB领域的图像-文字配对数据，优先采用高质量样本以实现视觉特征与标准化缺陷术语的对齐。第二阶段整合约30万组多轮对话，采用统一模式确保跨场景输出格式严格一致。第三阶段采用约5千个可验证样本（如多选题、JSON坐标）的精简集，以实现基于奖励的优化。表~\ref{table1}详细列出了各阶段的具体构成与数据量。所有阶段均采用统一的分类体系和标注规范，以最大限度减少分布偏移。
As illustrated in Fig.~\ref{fig5}, the constructed dataset is stratified into three subsets spanning multiple imaging modalities, organized by training stages, task families, and fine-grained question types. The first stage comprises approximately 100k caption pairs from general and PCB domains, prioritizing high-quality samples to align visual features with standardized defect terminology. The second stage incorporates roughly 300k multi-turn dialogues under a unified schema, ensuring strict output formatting across diverse scenarios. The third stage uses a condensed set of approximately 5k verifiable samples (e.g., MCQ, JSON coordinates) to enable reward-based optimization. The detailed data split is provided in the Appendix.
%Table~\ref{table1} details the specific composition and volume of each stage. %Consistent taxonomies and annotation conventions are applied throughout all stages to minimize distribution shifts.
\section{PCB-GPT}
%通用型超大规模语言模型在PCB检测中常因缺乏领域先验知识且定位精度要求极高而表现欠佳。为弥补这一缺陷，我们提出PCB-GPT模型。该方案引入定制化课程学习策略，通过模拟领域专家的认知成长路径，使模型能力从基础识别逐步提升至复杂推理阶段。
General-purpose MLLMs often falter in PCB inspection due to the lack of domain-specific priors and the high demand for localization precision. To bridge this gap, we propose PCB-GPT. We introduce a curriculum learning strategy tailored to mimic the cognitive growth of a domain expert, progressively enhancing the model's capabilities from basic recognition to complex reasoning.
\subsection{Three-stage curriculum}
Specifically, this strategy unfolds in three phases to endow PCB-GPT with: (i) PCB-specific visual semantics, (ii) instruction-following and structured generation, and (iii) verifiable decision-making ability via RL. We use Qwen2.5-VL-7B-Instruct as the base model and perform stage-wise optimization with LoRA~\cite{lora} while keeping the backbone frozen.
\paragraph{Stage-1: Concept Alignment}
%我们首先通过标题式概念数据将视觉编码器与PCB领域语义进行对齐。为在注入领域线索的同时保持通用视觉理解能力，我们冻结主模型，仅对自注意力机制的查询与值投影进行LoRA训练。此阶段主要提升了对PCB元件及缺陷概念的识别能力。
We first align the visual encoder with PCB domain semantics using caption-style concept data. To preserve general visual understanding while injecting domain cues, we freeze the backbone and train LoRA only on the Query and Value projections of self-attention. This stage mainly improves the recognition of PCB components and defect concepts.
\paragraph{Stage-2: Instruction Tuning}
% 接下来，我们对多模态问答数据进行监督式指令微调。我们冻结主模型架构，并对语言模型的所有线性层（输出头除外）应用LoRA技术。为促进显式推理与稳健解析，我们通过监督机制要求模型生成统一结构化的响应格式：
Next, we perform supervised instruction tuning on our multimodal QA data. We freeze the backbone and apply LoRA to all linear layers of the language model (excluding the output head). To encourage explicit reasoning and robust parsing, we supervise the model to generate responses in a unified structured format: \texttt{<think>\ldots</think><answer>\ldots</answer>}.
\paragraph{Stage-3: Reinforcement Learning with GRPO}
%为进一步提升可验证任务的性能，我们在第三阶段引入GRPO算法。针对每个提示，策略会采样一组候选响应，并从自动检查器处获取奖励。GRPO通过参考正则化目标函数执行组相对策略更新，同时我们额外加入严格格式奖励项以稳定结构化输出。在具体实现中，我们为每个提示采样$n{=}8$个响应，并对参考策略应用系数为0.01的KL正则化。
To further improve performance on verifiable tasks, we apply GRPO in Stage-3. For each prompt, the policy samples a group of candidate responses and receives rewards from automatic checkers. GRPO performs group-relative policy updates with a reference-regularized objective, and we additionally include a strict format bonus to stabilize structured outputs. In our implementation, we sample $n{=}8$ responses per prompt and apply KL regularization to the reference policy with coefficient 0.01.
\paragraph{Structured Reward for Verifiable Tasks}
%给定输入样本 x，模型生成输出 y，从中我们解析出答案片段 {a}(y)。对于具有基准答案 g 的可验证任务，我们定义：
Given an input sample $x$, the model generates an output $y$, from which we parse an answer segment $\hat{a}(y)$. For verifiable tasks with ground-truth $g$, we define:
\begin{equation}
R(x,y,g)
= w_T \cdot \mathrm{Acc}\,\big(\hat{a}(y), g\big)
+ w_F \cdot \mathbf{1}\!\left\{ y \in \mathcal{F} \right\},
\end{equation}
where $\mathrm{Acc}(\hat{a}(y), g)\in[0,1]$ is the automatic correctness score (e.g., exact match for MCQ, and IoU-based matching F1 for BBOX), and $\mathbf{1}\{y\in\mathcal{F}\}$ equals $1$ iff $y$ strictly follows the structured output format defined in Stage-2, otherwise $0$. We clip $R$ to $[0,1]$ and set $w_T=0.9$, $w_F=0.1$ in all experiments.
\subsection{Implementation details}
%所有阶段均在八块NVIDIA A100-40GB GPU上进行训练。第一阶段和第二阶段采用基于Swift~\cite{swift}的监督式微调，而第三阶段则利用Verl~\cite{verl}进行GRPO~\cite{grpo}策略优化。推理与评估通过vLLM~\cite{vllm}完成。表~\ref{table2}汇总了各阶段的学习率、序列长度及LoRA配置。第三阶段采用GRPO算法，每个提示词采样$n{=}8$个响应，KL系数设为0.01。
All stages were trained on eight NVIDIA A100-40GB GPUs. Stage 1-2 used Swift~\cite{swift} for supervised fine-tuning, while Stage 3 utilized Verl~\cite{verl} for GRPO~\cite{grpo} policy optimization. Inference and evaluation were conducted using vLLM~\cite{vllm}. %Table~\ref{table2} summarizes the learning rates, sequence lengths, and LoRA configurations for each stage. For Stage 3, we use GRPO with $n{=}8$ sampled responses per prompt and a KL coefficient of 0.01.
\section{Experiments}
\subsection{Overall Results on UniPCB Benchmark}
\paragraph{Experimental setup}
% On the UniPCB benchmark, we systematically evaluated multiple types of multimodal models. To ensure fair comparison, all models used a unified input prompt template and consistent decoding settings for inference. We compared the performance of numerous models, including IAD MLLMs, open-source MLLMs, and commercial models. Furthermore, for some data in unlabeled scenarios, we provided dual-image input 'normal image \& predicted image' to guide the model to complete defect detection through comparison. Apart from this setting, all other tasks used single-image input.
%我们在UniPCB基准测试上系统性地评估了多种模态模型，涵盖了IAD超大规模语言模型、开源超大规模语言模型及商用超大规模语言模型。为确保公平比较，所有评估均采用统一的提示模板和一致的解码设置。此外，针对无标签场景，我们采用双图像输入策略——将正常参考图像与查询图像配对，通过对比实现缺陷检测。除该特殊设置外，其余任务均采用单图像输入。
We systematically evaluated a diverse array of multimodal models on the UniPCB benchmark, covering IAD MLLMs, open-source MLLMs, and commercial MLLMs. To ensure fair comparison, a unified prompt template and consistent decoding settings were employed across all evaluations. Furthermore, for unlabeled scenarios, we adopted a dual-image input strategy—pairing a normal reference with the query image to facilitate defect detection via comparison. Apart from this specific setting, all other tasks utilized single-image inputs.
\paragraph{Main results}
Table~\ref{table3} presents the overall quantitative results on the UniPCB Benchmark. In general, commercial models maintain a competitive edge in semantic consistency and discriminative capabilities. 
GPT-5 retains the lead in open-ended QA, while Gemini 2.5 Pro achieves the highest accuracy in closed-ended QA, reflecting the robustness of commercial models in text comprehension and option discrimination.
Among open-source models, PCB-GPT achieves the highest average score and demonstrates significant superiority in coordinate localization. This finding highlights that within PCB quality inspection scenarios, strong language generation and semantic alignment do not automatically translate into reliable spatial localization. It further reveals that mainstream MLLMs still have considerable room for improvement regarding fine-grained localization and structured output generation.
In contrast, MLLMs designed for IAD maintain decent performance on semantic similarity metrics but generally exhibit weaker localization capabilities, leading to a notable gap in overall scores compared to general-purpose MLLMs. This aligns with the task assumptions of anomaly detection methods, which focuses on coarse-grained decisions about whether an anomaly exists and roughly where it occurs. Our benchmark instead emphasizes fine-grained defect analysis and precise localization, exposing a pivotal gap in existing capabilities.
\subsection{Cross-Dataset Generalization}
%我们将评估扩展至外部PCB-Bank数据集~\cite{pcb-bank}以验证跨数据集泛化能力。如表4所示，在零样本设置下，AnomalyGPT展现出高准确率但低召回率。这反映出因数据失衡导致的对多数非缺陷类别的偏好。PCB-GPT获得更高的F1分数，表明其对实际缺陷具有更强的敏感性。在单次学习场景下，竞争模型在召回率急剧提升的同时精度下降，表明在提供视觉参考时存在缺陷预测过度现象。相比之下，PCB-GPT通过利用上下文信息同时提升了准确率和精度。
We extended the evaluation to the external PCB-Bank dataset~\cite{pcb-bank} to verify cross-dataset generalization. As shown in Table~\ref{table4}, in the zero-shot setting, AnomalyGPT exhibits high accuracy but low recall. This reflects a bias toward the majority non-defect class due to data imbalance. PCB-GPT achieves a higher F1-score, indicating better sensitivity to actual defects. In the one-shot setting, competitor models show a sharp increase in recall paired with a drop in precision, implying an over-prediction of defects when provided with visual references. In contrast, PCB-GPT utilizes the in-context information to improve both accuracy and precision.
\begin{figure*}[!htbp]
    \centering
    \includegraphics[width=1\linewidth]{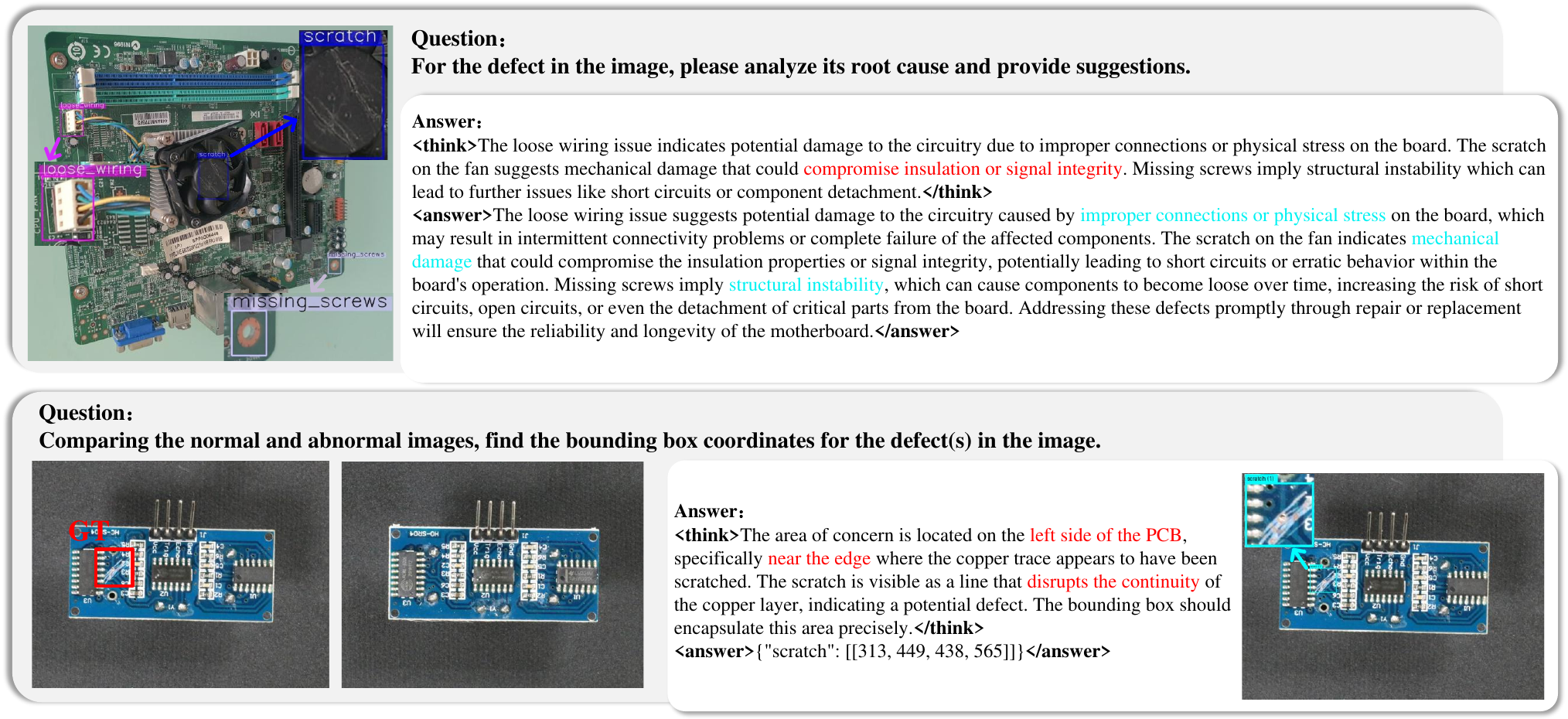}
    \caption{Qualitative examples of PCB-GPT on the UniPCB benchmark. Key logical evidence within the \texttt{<think>} block is highlighted in red, which explicitly guides the model to generate the final correct answers (highlighted in blue).}
%Qualitative examples of PCB-GPT on the UniPCB benchmark. The Chain-of-Thought (CoT) mechanism facilitates interpretable defect analysis and precise localization. Key logical evidence within the <think> block is highlighted in red, which explicitly guides the model to generate the final correct answers (highlighted in blue). This visualization demonstrates how the model grounds its structured outputs in specific visual characteristics and spatial contexts.
\label{fig6}
\end{figure*}
\begin{table}[t]
\centering
\footnotesize
\resizebox{\columnwidth}{!}{%
\begin{tabular}{l|cccc|cccc}
    \toprule
    Model & \multicolumn{4}{c|}{0-shot} & \multicolumn{4}{c}{1-shot} \\
    \cmidrule(lr){2-5}\cmidrule(lr){6-9}
     & Acc & Prec & Rec & F1 & Acc & Prec & Rec & F1 \\
    \midrule
    GPT5-Main & \underline{65.1} & 39.9 & 28.2 & 33.0 & \underline{48.3} & \underline{35.6} & \underline{85.7} & \underline{50.3} \\
    AnomalyGPT & \textbf{70.7} & \textbf{46.0} & \underline{44.6} & \underline{45.3} & 42.3 & 30.6 & \textbf{88.2} & 45.5 \\
    \textbf{PCB-GPT (Ours)} & 63.8 & \underline{40.6} & \textbf{72.1} & \textbf{52.1} & \textbf{77.8} & \textbf{50.6} & 76.0 & \textbf{61.0} \\
    \bottomrule
\end{tabular}}
\caption{Evaluation results on the PCB-Bank dataset. We report Accuracy (Acc), Precision (Prec), Recall (Rec), and F1-score under 0-shot and 1-shot settings.}
\label{table4}
\end{table}
%1.检查其他图和表引入是否得当
%2.前文部分内容需要指向后续附录
\subsection{Ablation Studies and Analysis}
\begin{table}[t]
\centering
\small
\setlength{\tabcolsep}{5pt}
\renewcommand{\arraystretch}{1.05}
\begin{adjustbox}{center}
\begin{tabular}{l|c|c|c|c|c}
\toprule
Model & CoT & Acc & F1 & OQA Score & Average \\
\midrule
Qwen2.5-VL        & \xmark & 48.7 & 22.3 & 59.9 & 43.6 \\
+BaseData        & \xmark & 60.1 & 30.0 & 66.1 & 52.1 \\
+Stage1          & \xmark & 57.3 & 15.2 & 62.8 & 45.1 \\
+Stage1,2        & \xmark & 57.1 & 22.7 & \underline{69.3} & 49.7 \\
+Stage1,2,3      & \xmark & \underline{65.6} & 27.3 & \textbf{69.8} & 54.2 \\
+Stage1,2        & \cmark & 63.3 & \underline{38.9} & 67.0 & \underline{56.4} \\
+Stage1,2,3      & \cmark & \textbf{69.5} & \textbf{51.1} & 69.0 & \textbf{63.2} \\
\bottomrule
\end{tabular}
\end{adjustbox}
\caption{Ablation on Qwen2.5-VL-7B. BaseData refers to unpartitioned and uncurated generated data.}
%CoT indicates chain-of-thought supervision
\label{table5}
\end{table}
%对Qwen2.5-VL-7B模型进行了分阶段训练与数据策略切除实验，结果详见表~\ref{table5}。即使不采用分阶段训练，纳入原始基础数据仍能带来稳定提升，表明更广阔的领域覆盖能增强通用识别与问答能力。然而定位指标的改进仍显有限且未达最优，这表明精确定位更多依赖后续指令对齐与结构化输出约束，而非单纯的数据积累。
Phased training and data strategy ablations were conducted on Qwen2.5-VL-7B, with results detailed in Table~\ref{table5}. Even without phased training, incorporating raw base data yields stable gains, suggesting that broader domain coverage enhances general recognition and QA capabilities. However, improvements in localization metrics remain limited and suboptimal. This implies that precise localization depends more on subsequent instruction alignment and structured output constraints than on mere data accumulation.
%我们进一步比较了不同训练阶段的贡献。概念对齐阶段主要提升了语义指标，但对整体性能影响有限，表明仅补充PCB领域知识不足以应对复杂的质量检测任务。引入指令微调后所有指标均获提升，证明指令监督能有效将知识转化为任务导向型输出。融入强化学习后性能进一步增强，尤其体现在输出一致性和可验证性方面。这证实了可验证样本衍生的奖励信号有助于稳定可精确评估任务的优化过程，从而增强系统鲁棒性。
We further compared the contributions of different training stages. The concept alignment stage mainly improved semantic metrics but had limited impact on overall performance, indicating that supplementing PCB domain knowledge alone is insufficient for complex quality inspection. The introduction of instruction fine-tuning improved all metrics, demonstrating that instruction supervision effectively translates knowledge into task-oriented outputs. Incorporating reinforcement learning further enhanced performance, particularly in output consistency and verifiability. This confirms that reward signals derived from verifiable samples help stabilize optimization for precisely evaluable tasks, thereby boosting robustness.
%此外，我们还考察了链式推理（CoT）监督的作用。采用CoT训练的模型取得了更优异的成绩，这并非源于文本长度的增加，而是得益于“先推理后作答”的显式约束。该约束迫使模型更有效地利用现有信息进行决策，从而提升了坐标等结构化输出的可靠性。相反，未采用CoT的模型在开放式问答任务中略占优势，这可能是因为省略推理过程减少了冗余，从而生成更简洁的回答。但这种优势仅限于开放式任务，对整体性能的提升作用有限。
Additionally, we examined the role of Chain-of-Thought (CoT) supervision. Models trained with CoT achieved superior scores, attributed not to increased text length but to the explicit 'reasoning before answering' constraint. This compels the model to better utilize existing information for decision-making, enhancing the reliability of structured outputs like coordinates. Conversely, models without CoT showed a slight advantage in open-ended QA, likely because omitting the reasoning process reduces redundancy, yielding more concise responses. However, this advantage is restricted to open-ended tasks and offers limited overall performance improvement.
\subsection{Qualitative Analysis}
Fig.~\ref{fig6} visualizes the inference process, where the CoT mechanism bridges visual perception and structured output. In the defect analysis case, the model explicitly derives potential risks from visual symptoms (e.g., 'loose wiring'), demonstrating strong interpretability. For localization, the \texttt{<think>} block explicitly identifies spatial cues (e.g., 'left side') prior to predicting coordinates. This indicates that the model learns to align visual features with spatial positions, ensuring the outputs are supported by image content.
\section{Conclusion}
%在本研究中，我们构建了系统化的数据构建流程，并开发了UniPCB——首个覆盖BPCB与PCBA场景的统一PCB质量检测基准。基于UniPCB，我们采用三阶段课程训练PCB-GPT模型以弥合领域差距，实现精细化定位与可解释推理。后续工作将扩展UniPCB的真实生产数据集，以提升工业检测的可靠性。
% In this work, we develop a systematic data construction pipeline and build UniPCB, the first unified benchmark for PCB quality inspection across BPCB and PCBA scenarios. Based on UniPCB, we train PCB-GPT with a three-stage curriculum to bridge the domain gap, enabling fine-grained localization and interpretable reasoning. Future work will expand UniPCB with real-world production data for more reliable industrial inspection.
UniPCB is a large-scale, bilingual benchmark unifying BPCB and PCBA levels. Structured into three progressive scenarios with 23k QA pairs, it evaluates robustness under varying information density. We propose PCB-GPT, a specialized assistant trained via a three-stage curriculum spanning concept alignment, instruction tuning, and reinforcement learning. This progressive paradigm bridges the domain gap, endowing the model with fine-grained localization and interpretable reasoning absent in general-purpose MLLMs. Future work will focus on scaling UniPCB with production data to advance industrial intelligence.
%在本研究中，我们提出一套系统化的数据构建管道，用于创建UniPCB——首个覆盖多样化BPCB和PCBA场景的PCB质量检测统一基准。基于此基础，我们提出PCB-GPT——通过创新的三阶段课程管道（涵盖概念对齐、指令微调与强化学习）训练的专用助手。这种渐进式范式有效弥合了领域鸿沟，赋予模型通用多语言模型普遍缺乏的精细定位与可解释推理能力。后续工作将聚焦于利用实际生产数据扩展UniPCB，以进一步推动可靠的工业智能发展。

% \appendix
% \section*{Ethical Statement}
% There are no ethical issues.
% \section*{Acknowledgments}
% \clearpage
\bibliographystyle{named}
\bibliography{main}



\section*{Supplementary material}

# UniPCB: A Unified Vision-Language Benchmark for Open-Ended PCB Quality Inspection

## Appendix

## A Data Curation and Splits

### A.1 Data sources and coverage

To maximize the diversity and representativeness of UniPCB, we first conducted a broad survey of PCB quality-inspection datasets spanning both academic benchmarks and industrial collections. This survey serves two purposes: (i) to systematically cover variations in *manufacturing stages* (bare boards vs. assembled boards), *imaging modalities* (optical to X-ray), and *inspection targets* (defects vs. components); and (ii) to provide a comprehensive reference for understanding the current data landscape, including common gaps in accessibility, annotation granularity, and label consistency.

Table 1 summarizes the resulting catalog of *raw candidate datasets* identified during this preliminary phase. Importantly, the table intentionally includes datasets with different accessibility statuses (e.g., private or closed-source industrial datasets). While such datasets are not used to construct the released benchmark, listing them helps contextualize the breadth of real-world inspection data and clarifies which parts of the landscape remain difficult to reproduce publicly.

**Field conventions in Table 1.** Each source is described using the following attributes:

- **PCB:** Manufacturing stage of the board. *BPCB* denotes bare printed circuit boards (unpopulated), whereas *PCBA* denotes assembled boards populated with electronic components.

- **Mod.:** Imaging modality / acquisition method. We use *RGB* for standard optical images, *AOI* for automated optical inspection systems, *Line-Scan* for linear scanning setups, *Ultrasonic* for ultrasonic imaging, and *AXI* for automated X-ray inspection. When a dataset is collected from in-the-wild or unconstrained environments (as opposed to controlled AOI/AXI pipelines), we denote it as *Real* to emphasize the capture condition and domain shift.

- **Target:** Primary inspection focus. *Defect* indicates anomaly/defect-centric inspection; *Component* focuses on component presence, type, or placement; and *Mix* contains both defects and components.

- **Statistics (#Lab. & #Img.):** **#Lab.** reports the number of categories *after mapping* the original labels to our unified taxonomy, and **#Img.** reports the original number of

images provided by the source.

- **Link:** Public URL or repository entry point for accessing the dataset when available.

**Reproducibility note** Although Table 1 includes private and closed-source datasets to reflect the full scope of PCB inspection data, the UniPCB benchmark is constructed strictly from the high-quality *open-source* subset. This design choice ensures that UniPCB can be independently reproduced, audited, and extended by the community.

### A.2 Preprocessing and Deduplication

**Annotation normalization**

Given the heterogeneity of source data, we harmonize diverse annotation formats—ranging from segmentation masks to normalized relative coordinates—into a unified axis-aligned bounding box representation. For sources with segmentation masks, we binarize the input and compute the tightest enclosing box derived from the min/max variance of foreground pixels. For annotations provided in relative formats, we project them into absolute pixel space, enforcing rigorous sanitization to clip boundaries, prune degenerate instances, and ensure spatial ordering ($x_1 < x_2, y_1 < y_2$). We adopt absolute pixel coordinates as the exclusive standard to mitigate ambiguity and guarantee compatibility across diverse model architectures (e.g., Qwen2.5 [Bai *et al.*, 2025], which is engineered for absolute inputs).

**Global deduplication**

To prevent data leakage arising from overlapping tiling, multi-pass captures, or partial re-releases, we execute a global deduplication protocol before any dataset splitting. The pipeline employs a two-level strategy: first removing byte-identical instances via file hashing, followed by perceptual hashing (pHash) to identify near-duplicates caused by minor resizing or compression artifacts. When a duplicate group is detected, we enforce a strict retention hierarchy that prioritizes evaluation integrity: Benchmark > Training. This ensures that visually identical regions—whether full images or overlapping crops—never cross-contaminate the training and evaluation sets.

### A.3 Train–Benchmark Isolation

UniPCB is designed to benchmark generalization under realistic inspection constraints. We therefore enforce strict iso-

| Dataset | PCB | Mod. | Target | #Lab. | #Img. | Link |
|---|---|---|---|---|---|---|
| HRIPCB [Huang *et al.*, 2020] | BPCB | RGB | Defect | 6 | 1386 | pkusz.edu.cn/ |
| HRIPCB-Augmented [Ding *et al.*, 2019] | BPCB | RGB | Defect | 6 | 10668 | github.com/ |
| DeepPCB [Tang *et al.*, 2019] | BPCB | Line-Scan | Defect | 6 | 3000 | github.com/ |
| PCB-AoI | PCBA | AOI | Defect | 1 | 1211 | kaggle.com/ |
| PCBA-DET [Shen *et al.*, 2024] | PCBA | Real | Defect | 8 | 4601 | github.com/ |
| Solder Joint Dataset [Ulger *et al.*, 2023] | PCBA | Real | Defect | 5 | 3390 | github.com/ |
| Dataset-PCB [Ramirez-Farfan and Quiroz-Martinez, 2021] | PCBA | Real | Defect | 2 | 3196 | github.com/ |
| DsPCBSD+ [Lei *et al.*, 2022] | BPCB | AOI | Defect | 9 | 10259 | github.com/ |
| PCB-Defect-Detection-Image-Registration [Tanthirige, 2022] | BPCB | Line-Scan | Defect | 6 | 20 | github.com/ |
| Defects Dataset | PCBA | Real | Defect | 5 | 484 | roboflow.com/ |
| PCB-Defect-Detection | PCBA | RGB | Defect | 6 | 898 | roboflow.com/ |
| Mono_PCB Dataset | PCBA | RGB | Defect | 16 | 248 | roboflow.com/ |
| Mixed PCB defect [Ancha *et al.*, 2024] | BPCB | RGB | Defect | 6 | 1741 | mendeley.com/ |
| Bangla_PCB_yolo | BPCB | RGB | Defect | 6 | 1196 | kaggle.com/ |
| MRC-DETR [Cao *et al.*, 2025] | BPCB | AOI | Defect | 3 | 800 | github.com/ |
| U-PCBD [Lou *et al.*, 2023] | BPCB | Ultrasonic | Defect | 5 | 4320 | iiplab.net/ |
| FICS [Lu *et al.*, 2020] | PCBA | RGB | Mix | 31 | 9912 | trust-hub.org/ |
| VisA (PCB) [Zou *et al.*, 2022] | PCBA | RGB | Mix | 10 | 4416 | github.com/ |
| PCB-Bank [Yao *et al.*, 2024] | PCBA | RGB | Mix | 11 | 2333 | github.com/ |
| PCB-Resistor-Defect-Dataset [Lv *et al.*, 2024] | PCBA | RGB | Mix | 11 | 261399 | github.com/ |
| PCB_Datasets-main | PCBA | AOI | Mix | 8 | 4748 | github.com/ |
| PCB AD | PCBA | RGB | Mix | 5 | 690 | kaggle.com/ |
| Multiple Datasets on PCB Defects | BPCB | AOI | Mix | 2 | 18493 | kaggle.com/ |
| MPI-PCB [Candido de Oliveira *et al.*, 2023] | PCBA | Real | Mix | 2 | 1797 | github.com/ |
| Micro-PCB Images | PCBA | RGB | Component | 13 | 8125 | kaggle.com/ |
| FPIC [Jessurun *et al.*, 2023] | PCBA | AOI | Component | 25 | 6260 | ece.ufl.edu/ |
| PCB oriented detection | PCBA | AoI | Component | 41 | 190 | kaggle.com/ |
| PCB Component Detection | PCBA | Real | Component | 9 | 1410 | ninja.com/ |
| PCB-Components-1495 | PCBA | Real | Component | 28 | 830 | kaggle.com/ |
| PCB-Component-Detection-CVM | BPCB | RGB | Component | 9 | 101 | roboflow.com/ |
| DSLR [Pramerdorfer and Kampel, 2015] | PCBA | RGB | Component | 4 | 748 | tuwien.ac.at/ |
| PCB-Vision [Arbash *et al.*, 2024] | PCBA | RGB | Component | 4 | 106 | github.com/ |
| Lite-On Dataset [Chen *et al.*, 2023] | PCBA | AOI | Mix | 8 | 12200 | Private |
| AXI_PCB_defect_detection [Zhang *et al.*, 2020] | PCBA | AXI | Mix | 4 | 32377 | Private |
| PCB-METAL [Mahalingam *et al.*, 2019] | PCBA | RGB | Component | 4 | 984 | Private |
| PCBMO [Li *et al.*, 2025] | PCBA | RGB | Component | 4 | 3773 | Private |
| PCBSDD [Liao *et al.*, 2021] | PCB | RGB | Component | 6 | 19029 | Private |

Table 1: Comprehensive catalog of raw PCB datasets surveyed for the construction of UniPCB. This inventory encompasses diverse PCB types (Bare/Assembled), imaging modalities (RGB, AOI, Line-Scan, X-ray, etc.), and accessibility statuses (Open-source vs. Private). Note that UniPCB selectively curates and standardizes high-quality samples from the open-source subset of these collections.

| Stage | Data Composition | Amount |
|---|---|---|
| Concept Alignment | **1. General image captions:** Pixmo [Deitke *et al.*, 2025]<br>**2. Defect image captions:** Defects Dataset, PCB-Defect-Detection, DsPCBSD+, PCBA-en, PCBAD, PCB-Bank, PCB-Resistor-Defect-Dataset-A<br>**3. Component image captions:** PCB-Components-1495-A, Micro-PCB Images | ∼100K |
| Instruction Tuning | **1. General multimodal:** LLaVA-1.5 Instruction [Liu *et al.*, 2023]<br>**2. Defect image multimodal:** Bangla_PCB_yolo, Mono_PCB dataset, MRC-DETR, HRIPCB, Mixed PCB defect, HRIPCB-Augmented, PCB_Datasets-main, Multiple Datasets on PCB Defects<br>**3. Component image multimodal:** PCB oriented detection, PCB-Components-1495-B | ∼300K |
| Reinforcement Learning | **1. Defect images:** Bangla_PCB, Mono_PCB, MRC-DETR, HRIPCB-Augmented<br>**2. Component images:** PCB-Component-Detection-CVM, PCB-Components-1495-B | ∼5K |
| Benchmark | **1. Defect images:** PCBA-DET, Dataset-PCB, PCB-AoI, DeepPCB, VisA(PCB), Solder Joint Dataset, PCB-Defect-Detection-Image-Registration<br>**2. Component images:** PCB-Resistor-Defect-Dataset-B, FPIC, PCB-component-detection | ∼23K |

Table 2: Stages, corresponding training datasets, and dataset amounts (some dataset names are abbreviated).



(a) English Subset

(b) Chinese Subset

Figure 1: Bilingual word clouds of the training dataset: (a) English and (b) Chinese high-frequency keywords. Dominant component and defect terms verify the corpus's domain specificity.



Figure 2: Distribution of image resolutions.

lation between (i) training data used for model development (including curriculum stages) and (ii) the UniPCB benchmark used for reporting results.

**Stage-wise curriculum data construction**

Table 2 summarizes the four data blocks used in our pipeline: concept-alignment data, instruction-tuning data, reinforce-ment learning (verifiable) data, and the benchmark set. We use dataset-level partitions whenever possible; for large sources, we create mutually exclusive partitions denoted by suffixes such as "-A" and "-B" and allocate them to different stages to avoid cross-stage contamination.

**Isolation strategies**

We apply three complementary strategies: (i) Source isolation: benchmark images are preferentially drawn from held-out datasets and held-out partitions. (ii) Cross-split dedu-plication: we run near-duplicate matching across all splits and remove any overlapping samples. (iii) Rule-level iso-lation: benchmark prompts/templates are designed indepen-dently from training templates to reduce the chance of mem-orizing surface forms.

**A.4 Dataset statistics**

To provide a comprehensive characterization of the dataset, Fig. 1 first visualizes the high-frequency keywords in the bilingual subsets, where the prominence of component terms (e.g., resistor", capacitor") and defect terms (e.g., "short cir-cuit") confirms that the training corpus contains rich domain-

| Category | ID | Name | Description | Impact / Function |
|----------|-----|------|-------------|-------------------|
| Structural | 1 | Bad Alignment | Misregistration shifts traces or pads. | Opens, shorts, or impedance mismatch. |
| | 2 | Hole Offset | Drill hole misaligned relative to pad. | Affects connectivity, insertion; risks opens. |
| | 3 | Crack | Substrate or component stress cracks. | Propagates to opens; undermines reliability. |
| | 4 | Physical Damage | Breakage or chipping from handling. | Compromises integrity; opens or shorts. |
| | 5 | Missing Hole | Required hole undrilled; process failure. | Electrical opens or insertion failure. |
| | 6 | Missing Pad | Pad missing; etching or adhesion error. | Prevents attachment; leads to opens. |
| | 7 | Mouse Bite | Semi-circular depressions along trace edge. | Reduced current capacity; risk of opens. |
| Soldering | 8 | Cold Solder | Dull, grainy surface; insufficient heat. | High resistance; intermittent electrical contact. |
| | 9 | Excessive Solder | Convex, bulbous joint exceeds pad. | Risks shorts; masks wetting issues. |
| | 10 | Lack of Solder | Low solder volume; insufficient fillet. | Weak joint; intermittent or open circuits. |
| | 11 | Pad Corrosion | Oxidized surface from poor storage. | Dewetting; bad contact or opens. |
| | 12 | Solder Bridge | Unintended connection between adjacent pads. | Short circuits; device burnout risk. |
| Component-Related | 13 | Connector_Damage | Cracked housing or bent pins; poor mating. | Interrupts power/signals; unstable communication. |
| | 14 | Pin Damage | Component leads bent, broken, or detached. | Causes poor contact, opens; functional failure. |
| | 15 | Missing Component | Component absent, not placed, or detached. | Incorrect functionality or complete module failure. |
| | 16 | Missing Switch | Switch physically absent from designated pads. | No user input; control logic fails. |
| | 17 | Component Shift | Misaligned component offset from pad center. | Unreliable connections, shorts, or instability. |
| Wiring connection | 18 | Open Circuit | Discontinuity in trace, pad, or joint. | Interrupts signals; causes malfunction or failure. |
| | 19 | Short Circuit | Unintended connection between isolated conductors. | High current; device burnout or failure. |
| | 20 | Spur | Copper projection extending from trace edge. | Bridges gaps; risks shorts, signal noise. |
| | 21 | Spurious Copper | Isolated copper in unintended areas. | Risks shorts, parasitic coupling, EMC. |
| Mechanical | 22 | Scratch | Abrasion on copper, pad, or mask. | Exposes metal; risks opens, interference. |
| Contamination | 23 | Burnt | Localized charring or blistering from heat. | Irreversible structural damage; fire safety hazard. |
| | 24 | Contamination | Surface residues: flux, oil, fingerprints, or chemicals. | Causes leakage, corrosion, or poor solder wetting. |
| | 25 | Glue Contamination | SMT adhesive residue contaminating the solder pads. | Prevents wetting; leads to dry, unstable joints. |
| | 26 | Adhesive Bubbling | Uneven dispensing; trapped air bubbles. | Component shifting, tombstoning, adhesion failure. |
| Foreign-Matter | 27 | Foreign Material | Dust, metal chips, or debris. | Shorts, corrosion; reduced reliability. |
| Control-Computing | 1 | IC | Microchip (SOP/QFP/BGA) for logic or processing. | Core logic; failure disables major functions. |
| | 2 | Microcontroller Board | MCU/SoC-based board (Arduino, Raspberry Pi). | Central controller; failure affects overall operation. |
| Interconnect | 3 | Connector | Plug/mate interconnect for cables or modules. | Carries power/signals; disconnect causes failure. |
| | 4 | Socket | Receptacle for detachable connections. | Eases maintenance; poor contact drops signal. |
| | 5 | Terminal Block | Screw/clamp terminal for external wire connection. | Robust wiring; loosening causes opens or resistance. |
| | 6 | Jumper/Header | Pin header with configuration shunt. | Sets mode; missing shunt alters state. |
| Passive | 7 | Capacitor | Stores energy; filters, couples, or decouples. | Stabilizes power; failure causes noise or resets. |
| | 8 | Inductor | Magnetic storage for filtering or power conversion. | EMI filtering; failure hurts efficiency or stability. |
| | 9 | Resistor | Limits current, divides voltage, or biases. | Sets operating points; drift alters circuit limits. |
| | 10 | Crystal/Oscillator | Quartz/active oscillator generating system clock. | Provides timing; failure stops MCU logic execution. |
| | 11 | Transformer | Voltage conversion and isolation. | Power failure; circuit damage. |
| Display-Interaction | 12 | LED | Light-emitting diode for status indication. | Visual feedback; failure reduces system observability. |
| | 13 | LCD | Display module for complex visual output. | User interface; failure blocks data display. |
| | 14 | Switch | Mechanical switch for power or mode control. | User input; failure breaks interaction or logic. |
| | 15 | Buzzer | Audio signaling device (piezo or magnetic). | Audible alarm; failure causes loss of warning. |
| | 16 | Potentiometer | Variable resistor for manual calibration/adjustment. | Analog input; wear causes noise or drift. |
| Active | 17 | Diode | One-way device for rectification or protection. | Prevents reverse current; failure risks circuit damage. |
| | 18 | Transistor | Semiconductor device for switching or amplification. | Signal driver; failure disables loads or overheats. |
| | 19 | Relay | Electromechanical switch; high-power isolation. | Mechanical failure; switching stops. |
| Power-Protection | 20 | Battery | On-board energy source or backup power. | Power supply; failure causes shutdown or data loss. |
| | 21 | Fuse | Protection device; melts at excessive current. | Prevents fire; blown fuse cuts circuit power. |
| | 22 | Voltage Regulator | Regulates input to stable voltage rails. | Stabilizes voltage; failure causes system crash. |

Table 3: Unified taxonomy of PCB defect and component instances. We provide detailed attributes (Category, ID, Name, Description, Impact/Function, etc. ) as textual priors. The IDs link directly to the numerical indices in Fig. 4 and Fig. 5.

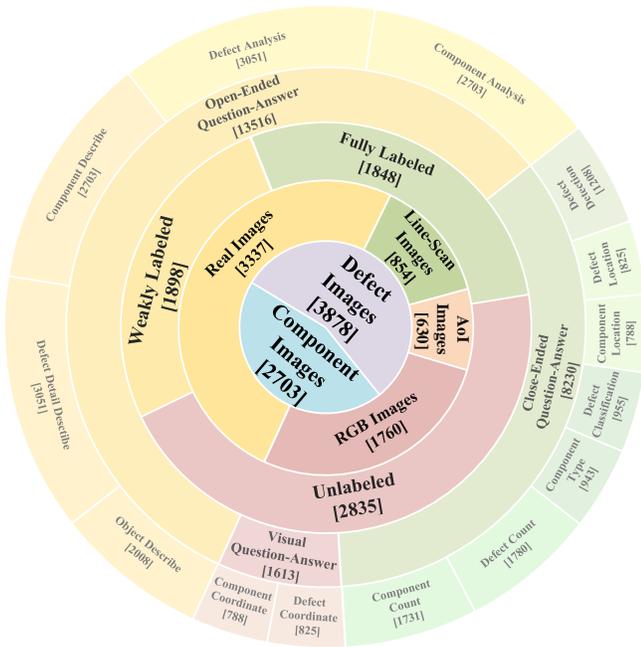

Figure 3: Statistical distribution of the UniPCB Benchmark. The sunburst chart visualizes the hierarchical organization of the benchmark across five nested levels.

specific semantics rather than generic captions. Complementing this textual depth, Fig. 2 illustrates the extensive distribution of image resolutions across Stages 1-4. The dataset spans diverse aspect ratios—from square formats (1:1) to standard photographic (4:3) and wide industrial ratios (16:9)—which prevents the model from overfitting to specific dimensions and necessitates our robust, resolution-agnostic output design. Finally, Fig. 3 details the hierarchical organization of UniPCB Benchmark, flowing from instance counts through diverse modalities (RGB, AOI, Real) and scenarios to specific QA types and 14 distinct inspection subtasks; this design ensures the benchmark jointly varies difficulty while maintaining a unified set of task definitions.

## B Unified Taxonomy and Task Schema

### B.1 Taxonomy: Definitions and Label Mapping

To establish a consistent and extensible framework, UniPCB consolidates heterogeneous inspection labels into the canonical taxonomy listed in Table 3. Each entry includes a definition, visual symptoms, and functional impact, acting as domain priors for model reasoning. Structurally, we organize the taxonomy hierarchically: defect classes are grouped into high-level families (e.g., structural, soldering, mechanical, contamination) to reflect manufacturing failure modes, while component classes are categorized by functional roles (e.g., control, interconnect, passive, power) to mitigate inter-class similarity in dense layouts. In implementation, we align raw labels from source datasets to these canonical definitions, merging synonymous terms for identical physical failures and discarding samples with underspecified evidence. Fig. 4 and

Fig. 5 visualize the finalized label space, presenting representative exemplars for each category.

### B.2 Task Formats and Output Constraints

UniPCB follows practical inspection workflows and defines 14 subtasks spanning detection, counting, classification, localization, description, defect analysis, and re-inspection suggestions. To evaluate robustness under different information density, we provide three scenarios: fully labeled (category and box), weakly labeled (box only), and unlabeled (no annotations).

**Answer types**

The benchmark incorporates three distinct response formats to evaluate diverse capabilities. Open-ended QA targets high-level description and analysis through free-form generation, whereas Closed-ended QA assesses discrimination ability via multiple-choice or binary questions with fixed answers. Additionally, Visual Grounding demands precise spatial localization, requiring the model to output structured and verifiable coordinates.

**Prompts used for generation**

Fig. 6 illustrates the progressive prompts used to generate instruction data for our three-stage curriculum: Stage-1 emphasizes observation-centric, caption-like descriptions; Stage-2 introduces task-explicit instructions with deterministic ground-truth alignment for verifiable formats (MCQ and localization); Stage-3 produces omni-task instruction-style conversations with concise user-facing answers and without explicit reasoning tags. Fig. 7 shows the standardized benchmark prompt, which enforces single-answer questions, strict language consistency (English-only or Chinese-only per sample), and type-dependent format constraints (exact-match letters for MCQ and JSON fields for localization) to ensure robust automatic evaluation.

## C Quality Control and Supplements

### C.1 Dual-track QC and retrospective refinement

To ensure that UniPCB is both reliable and automatically evaluable, we enforce data integrity through a strict dual-track mechanism combining model-based automation with human expertise. The pipeline begins with automated verification, where a lightweight 7B-parameter MLLM is deployed to filter structural anomalies, guaranteeing JSON parsability and basic field completeness before human intervention. Samples clearing this threshold proceed to a hierarchical manual review, where experts validate the data against five logical dimensions: CoT logic consistency, coordinate validity, field completeness, scenario compliance, and JSON formatting. Finally, we establish a retrospective adjustment loop to drive continuous improvement, error patterns identified during manual inspection trigger iterative refinements of the upstream vision prompts and QA rules. Crucially, this mechanism strictly locks the logic of the generated Benchmark Set to ensure evaluation stability, while restricting continuous optimizations to the modifiable training branch.

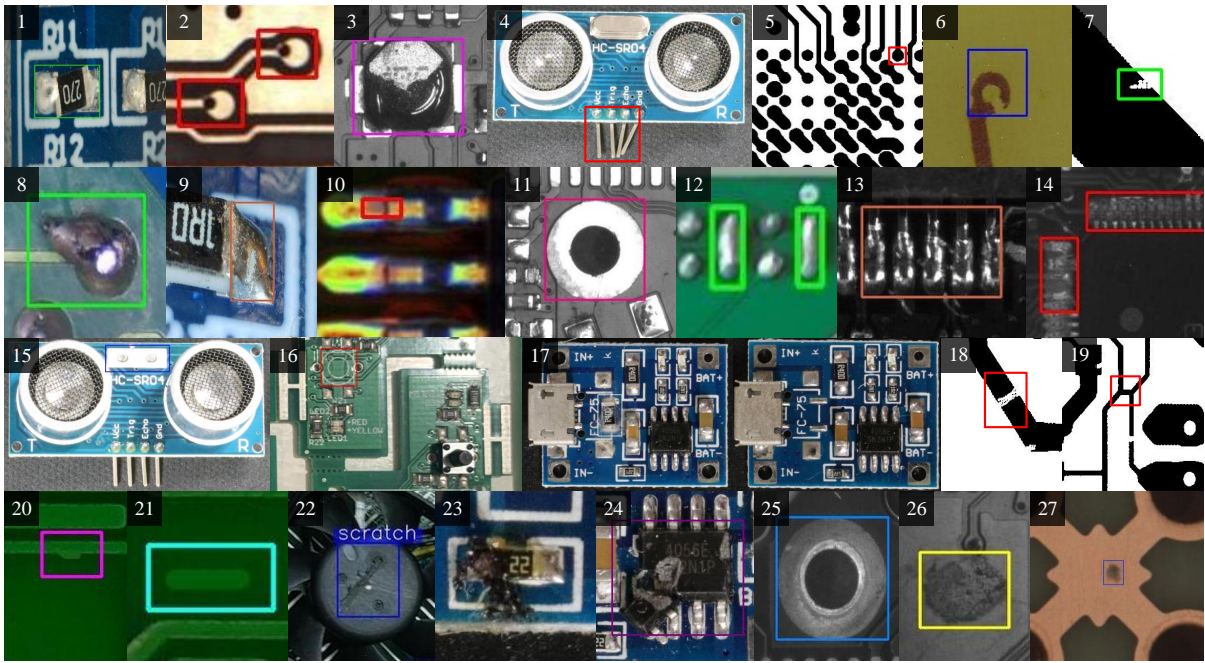

Figure 4: Visual examples of PCB defect instances. The indices in the top-left corners correspond to the IDs in Tab. 3.

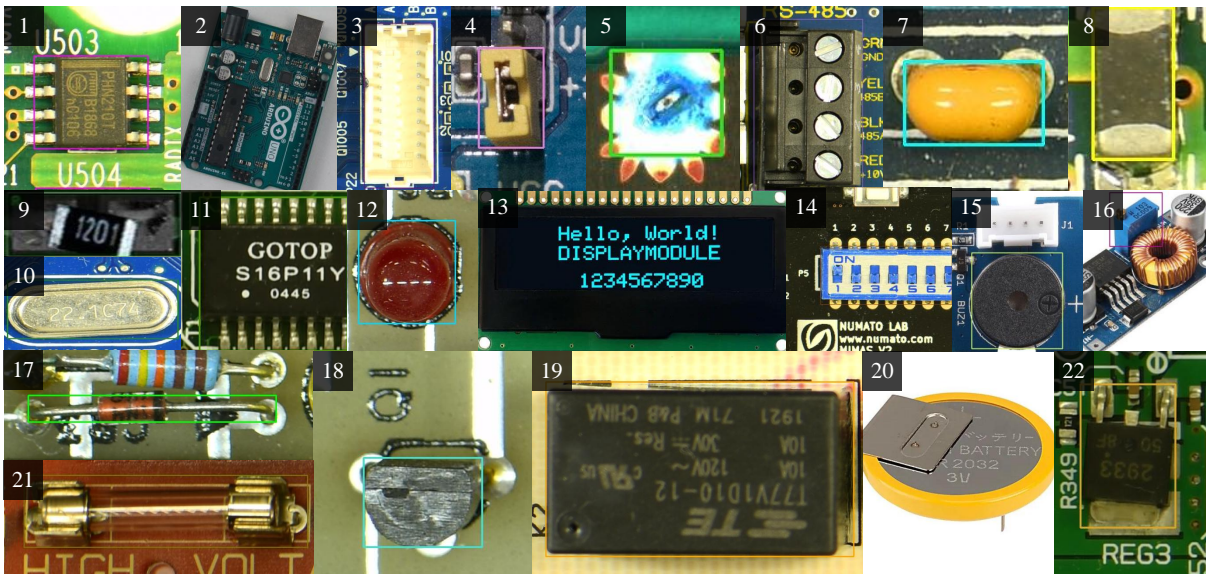

Figure 5: Visual examples of PCB component instances. The indices in the top-left corners correspond to the IDs in Tab. 3.

| Stage | LoRA Scope | Rank/$\alpha$ | Seq Len | LR |
|---|---|---|---|---|
| Stage1 | Attn Q/V | 32/64 | 4096 | 1e-4 |
| Stage2 | All linear | 64/128 | 6144 | 1e-4 |
| Stage3 | All linear | 32/64 | 4096 | 3e-6 |

Table 4: Training configurations across the three stages.

## C.2 Additional qualitative examples

Fig. 8–10 illustrate the critical performance gap between PCB-GPT and general-purpose baselines across three representative failure modes. In scenarios requiring semantic-spatial alignment, generalist models often exhibit a disconnect between verbal retrieval and visual grounding; for instance, while they may correctly identify a defect like a 'bent pin', their predicted bounding boxes frequently drift toward salient background features, indicating a fundamental failure in performing differential analysis. This limitation is exacerbated in crowded PCBA scenes with high inter-class similarity, where baselines exhibit universal failure—often suffering from spatial hallucination by generating boxes on empty traces (e.g., GPT-5), whereas PCB-GPT accurately isolates the target by effectively aligning fine-grained visual cues. Finally, regarding compositional reasoning for co-occurring defects (e.g., 'lack of solder' combined with 'solder bridge'), general models typically display a single-label bias that ignores secondary faults. In contrast, PCB-GPT demonstrates superior reasoning capabilities, aggregating multiple visual evidences to correctly identify compound failure modes rather than defaulting to partial answers.

## D Training Configuration

Table 4 summarizes the training configurations across the three curriculum stages, including LoRA scope, rank/$\alpha$, sequence length, and learning rate. In Stage-1, we restrict LoRA to attention Q/V projections to preserve general visual ability while injecting PCB semantics. In Stage-2, we apply LoRA to all linear layers to strengthen instruction following and structured generation. In Stage-3, we use a smaller learning rate for stable optimization on verifiable samples.

# # Prompt for Instruction Data Generation

- Role

You are an AI assistant specialized in PCB inspection, capable of perceiving visual content and providing professional, task-aware responses based on circuit board images.

- Overall Objective

Generate high-quality image–text instruction data for vision–language model training, with progressively stronger supervision on task specificity, reasoning, and instruction following.

## Stage-1: Caption-like Visual QA Prompt

- Purpose

To construct natural, observation-driven image–text question answering pairs that resemble human visual inquiries.

- Instructions

1. Generate caption-style questions inspired by the image content.
2. Questions must be natural and descriptive, avoiding task-like instructions.
3. Ground truth knowledge is hidden and may only be used to improve the answer quality.
4. Do not reveal defect types, quantities, locations, or annotations in the question.
5. When multiple QA pairs are generated, each should focus on a different visual aspect.

- Output Format

1. JSON object containing alternating user–assistant turns.
2. Each user turn must start with <image> followed by the question.

## Stage-2: Task-oriented Visual Instruction Prompt

- Purpose

To generate task-explicit visual instructions targeting specific inspection capabilities.

- Instructions

1. Each question must correspond to a predefined task type (e.g., detection, count, classification, location, analysis).
2. Questions are task-aware and image-grounded.
3. Explicit reasoning is required for selected tasks to guide visual inference.
4. Deterministic tasks must match the ground truth exactly.
5. Answers must follow the structured format:
    <think>Visual evidence and step-by-step reasoning.</think>
    <answer>Final task-specific answer.</answer>

- Constraints

1. Ground truth information must not appear in the question.
2. Reasoning must be visually grounded and avoid referencing labels, annotations, or ground truth terms.
3. Deterministic tasks require answers that strictly match the ground truth.

## Stage-3: Omni-task Instruction Conversation Prompt

- Purpose

To align the model with real-world instruction-following behavior across diverse visual tasks.

- Instructions

1. Support diverse task formats, including: QA, multiple-choice questions, yes/no questions, and structured outputs.
2. Questions must remain task-aware and image-grounded.
3. Responses should be concise, professional, and user-facing.
4. Do not include explicit reasoning tags (e.g., <think>).
5. Multi-turn conversations may combine multiple tasks over the same image.

- Output

1. Instruction-style conversations suitable for large-scale instruction tuning.
2. Clean answers without exposed internal reasoning.

Figure 6: The progressive prompts used to generate the three-stage curriculum instruction tuning data.

# # Prompt for Benchmark Construction

- Role

You are an AI assistant for constructing standardized PCB inspection benchmarks with unambiguous questions and automatically verifiable answers.

- Overall Objective

To generate evaluation-ready VQA and multiple-choice questions with fixed answers, controlled difficulty, and consistent language, enabling reliable model comparison.

- Instructions

1. Generate benchmark questions based on image content and dataset annotations.
2. Each question must have a single correct answer.
3. For multiple-choice questions:
   (1) Generate one correct option and multiple plausible distractors.
   (2) Distractors must be consistent in style and length while being clearly incorrect.
4. Enforce strict language consistency (Chinese or English only).
5. Output must be machine-readable and suitable for automatic scoring.

- Supported Question Types

1. Object description
2. Defect: detection, classification, count, location, coordinates, analysis
3. Component: type, count, location, description, coordinates, analysis
4. Comparative questions (e.g., OK vs. NG)

- Output Format

1. JSON with explicit fields for:
   (1) question
   (2) correct_option
   (3) options (if applicable)
2. Structured coordinate outputs when required.

Figure 7: The standardized prompt used for UniPCB benchmark construction.

**Question:**
Comparing the normal and abnormal images, Find the absolute bounding box coordinates for the bent defect(s) in the image.
Inside <answer>, output ONLY JSON coordinates: {"Category Name": [[x1,y1,x2,y2], ...], ...}.

**GPT-5:**
<answer> { "Bent Pin": [ [430, 690, 520, 840] ] } </answer>

**Gemini2.5-Pro:**
Based on the visual comparison between the normal and abnormal images, the defect is identified as a bent pin on the ultrasonic sensor module (HC-SR04).
Specifically, the leftmost pin (typically the Vcc pin) is bent significantly to the left compared to the straight alignment seen in the normal reference image.
<answer> {"bent_pin": [[465, 720, 535, 890]]} </answer>

**Claude-Sonnet 4.5:**
Looking at the two images, I can identify the bent defect by comparing the normal image with the abnormal image.
The bent defect is visible on one of the four pin connectors at the bottom of the module. In abnormal image, the rightmost pin appears to be bent compared to normal image where all pins are straight.
<answer> {"Bent": [[0.85, 0.82, 0.92, 0.95]]} </answer>

**Doubao-seed-v1.6:**
{"Bent Defect": [[375,730,403,804]]}

**Qwen3-Max:**
<answer>{"bent defect": [[380,715,416,804]]}</answer>

**PCB-GPT:**
<think>In the first image, there is a noticeable deformation on one of the pins at the bottom left corner, which aligns with the definition of a bent defect as it involves a pin that has been bent out of its original position. The second image does not show any such deformation, confirming the presence of a bent defect in the first image.</think>
<answer>{"bent": [[528, 790, 613, 884]]}</answer>

Figure 8: Visualization of reference-based defect localization. The example targets a "bent pin" defect on an ultrasonic sensor, utilizing a normal reference image and a query image. The text panel displays the internal <think> process preceding the final coordinate output.

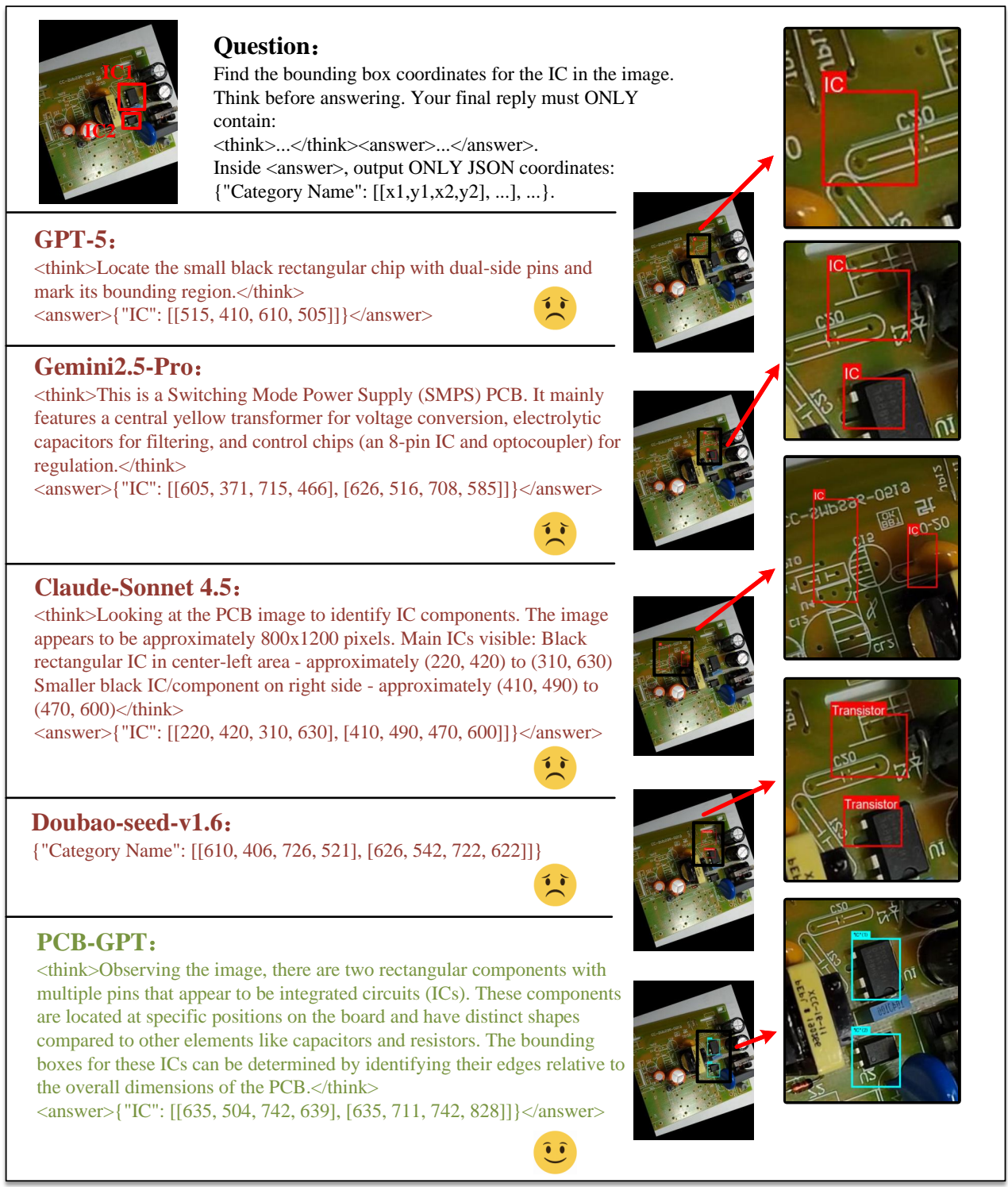

Figure 9: Visualization of dense component localization. The task requires identifying the bounding box of an 'IC' (Integrated Circuit).

**Question:**

Regarding the labeled target in the image, what is its type?

A. solder bridge.

B. component damaged.

C. lack of solder and open circuit.

D. lack of solder and solder bridge.

Select the option you think is correct and output only its uppercase letter.

Think before answering. Your final reply must ONLY contain:

<think>...</think><answer>...</answer>.

Inside <answer>, output ONLY one letter A/B/C/D.

**GT:**

<answer>A</answer>.

<think>The image shows multiple labeled regions: yellow boxes indicate insufficient solder (lack of solder) on through-holes, and green boxes at the bottom indicate solder bridges between pads. Therefore, the labeled targets include both lack of solder and solder bridge defects.</think>

<answer>D</answer>

**Gemini2.5-Pro:**

<think>The yellow boxes indicate a lack of solder on individual joints, while the green boxes highlight solder bridges connecting multiple pads; therefore, option D correctly identifies both defect types present in the image.<think>

<answer>D</answer>

**Claude-Sonnet 4.5:**

<think>Looking at this PCB image, several areas are highlighted with yellow boxes. The highlighted spots appear to be solder joints or vias. Examining them closely, particularly the two circular areas on the lower left, they seem to show insufficient solder coverage or missing solder. This type of defect would prevent proper electrical connection, resulting in an open circuit. I don't see evidence of solder bridges (unwanted connections between pads) or damaged components. The defect pattern is consistent with inadequate soldering leading to connectivity issues, making option C the most appropriate answer.</think>

<answer>C</answer>

**Doubao-seed-v1.6:**

A

**Qwen3-Max:**

<answer>C</answer>

**Grok:**

<think>The labeled targets (green and yellow boxes) point to several solder joints around components like the IC, resistors, and capacitors. Most appear to have insufficient solder fillet or barely any solder coverage on the pads, indicating poor wetting or missing solder material, which can lead to open circuits. There is no clear continuous solder blob connecting adjacent pads that would qualify as a solder bridge. No components show visible cracking, breaking, or obvious physical damage either. Therefore the primary defect visible in the labeled areas is lack of solder (and potential open circuit as a consequence), not bridging or component damage.</think>

<answer>C</answer>

**PCB-GPT:**

<think>Observing the highlighted areas on the PCB, there are visible signs of soldering issues such as insufficient solder coverage around the pads and pins, which can lead to poor electrical connections. The highlighted regions indicate potential solder bridges where solder has spread beyond intended connections, leading to unintended conductive paths between components that should be isolated.</think>

<answer>D</answer>

Figure 10: Visualization of multi-type defect classification. The task requires identifying co-existing defects ('lack of solder' and 'solder bridge') and selecting the corresponding compound option (D) from the multiple-choice list.


\end{document}